\pdfoutput=1
\let\ul\relax
\documentclass[11pt]{article}

\usepackage[final]{acl}

\usepackage{times}
\usepackage{latexsym}
\usepackage{bm}
\usepackage{comment}
\usepackage{xurl}

\usepackage{subcaption} 

\usepackage[T1]{fontenc}

\usepackage[utf8]{inputenc}

\usepackage{microtype}

\usepackage{inconsolata}
\usepackage{graphicx}
\usepackage{multirow}
\usepackage[normalem]{ulem}
\useunder{\uline}{\ul}{}

\usepackage{stfloats}
\usepackage{afterpage}

\usepackage{booktabs}
\usepackage{multirow}
\usepackage[normalem]{ulem}
\usepackage{makecell}
\usepackage{tabularx}
\usepackage{float}
\usepackage{balance}

\usepackage{kotex}

\usepackage{soul} 
\usepackage{makecell}

\usepackage{amsmath}
\usepackage{mathtools}
\usepackage{comment}
\usepackage{amssymb}
\usepackage{xspace}
\usepackage{adjustbox}

\usepackage{listings}
\usepackage{enumitem}
\usepackage{lscape}

\usepackage{tabularx} 
\usepackage[hang,flushmargin]{footmisc}

\usepackage{footmisc}

\newcommand{\sys}{Thunder-DeID\xspace}

\title{\sys: Accurate and Efficient De-identification Framework \\ for Korean Court Judgments}

\newcommand{\samethanks}[1][\value{footnote}]{\footnotemark[#1]}

\author{
Sungeun Hahm\thanks{Equal contribution.}$^{1}$\;
Heejin Kim\samethanks$^{1}$\;
Gyuseong Lee\samethanks$^{1}$\; 
Hyunji M. Park$^{1}$\;
Jaejin Lee$^{1,2}$ 
\\
$^{1}$Graduate School of Data Science, Dept. of Data Science, Seoul National University \\
$^{2}$College of Engineering, Dept. of Computer Science and Engineering, Seoul National University \\
\texttt{\{isungeuni, kheejin, ksnannaya, mhj233, jaejin\}@snu.ac.kr}
}

\begin{document}
	\maketitle
	\begin{abstract}
To ensure a balance between open access to justice and personal data protection, the South Korean judiciary mandates the de-identification of court judgments before they can be publicly disclosed. However, the current de-identification process is inadequate for handling court judgments at scale while adhering to strict legal requirements. 
Additionally, the legal definitions and categorizations of personal identifiers are vague and not well-suited for technical solutions. To tackle these challenges, we propose a de-identification framework called \sys, which aligns with relevant laws and practices. Specifically, we (i) construct and release the first Korean legal dataset containing annotated judgments along with corresponding lists of entity mentions, (ii) introduce a systematic categorization of Personally Identifiable Information (PII), and (iii) develop an end-to-end deep neural network (DNN)-based de-identification pipeline. Our experimental results demonstrate that our model achieves state-of-the-art performance in the de-identification of court judgments.
\end{abstract}

	\section{Introduction}

Generally, court proceedings are open and accessible to the public. It is one of the key democratic principles enshrined in the constitutions of many countries, including South Korea\footnote{Constitution of South Korea, Art. 109}. South Korea is one of the countries with more stringent conditions that cover a broader range of personal identifiers to be anonymized in the court setting. 


Before the publication of court decisions, the Korean National Court Administration uses both manual and automated de-identification methods throughout four stages of processing and review~\cite{judicial2021personaldata}. However, the current state of the de-identification procedure is not capable of handling court judgments at scale.

We want to address the following three problems of the current state of the de-identification procedure in South Korea. First, over-reliance on the manual method has been a major bottleneck, causing administrative strain and delaying publication of judgments. Public accessibility of judgments has been significantly low in South Korea, and the stagnant de-identification procedure is one of the reasons~\cite{nca2025courtpublication}.
Second, the automatic de-identification tool's performance is surprisingly low. From 2019 to 2025, their overall accuracy merely spans  8 to 15\%~\cite{congressional2019openpublication, nca2025courtpublication}. Finally, while existing law lays out the scope of de-identification, how personal identifiers are categorized and defined for administrative practice at the court is vague and especially unsuitable to be used for automated technical solutions. 


\label{sec:intro}

\begin{figure*}
    \centering
    \includegraphics[scale=0.82, width=\textwidth]{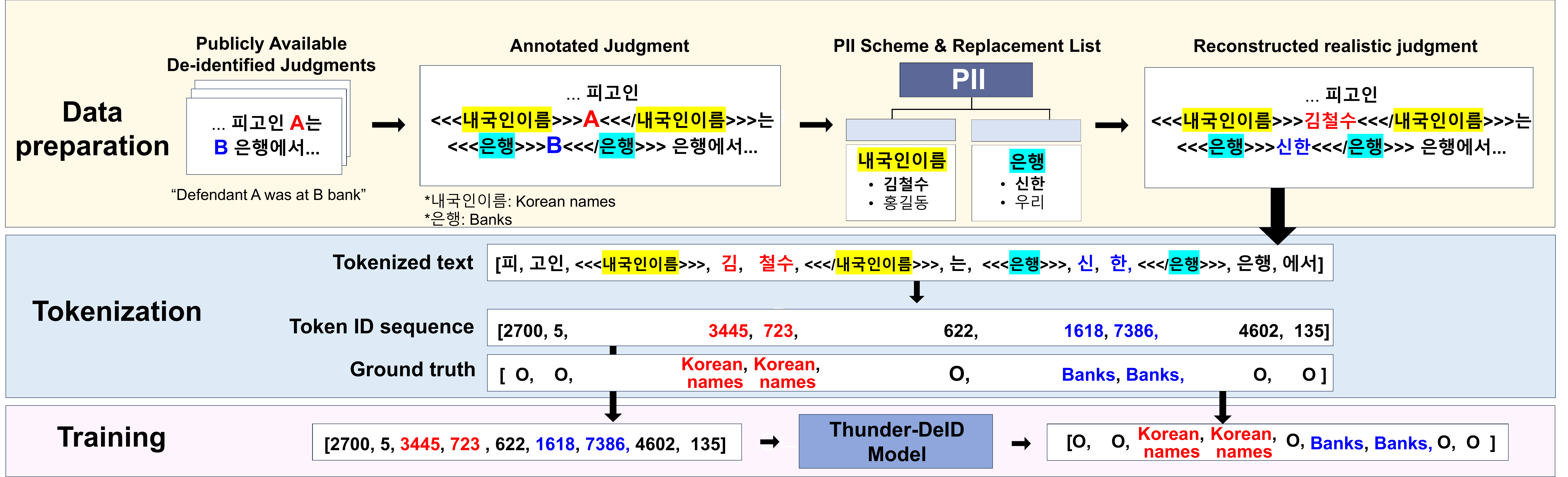}
       \vspace{-1.5\baselineskip}
    \caption{Overview of \sys.}
    \label{fig:overall}
\vspace{-3mm}
\end{figure*}

To overcome the above problems, this paper proposes \sys, a DNN- and NER-based framework, which improves the accuracy, efficiency, and consistency of de-identifying court judgments. Unlike a prompt-based approach using a large language model (LLM), which often alters the original sentence structure in the process of deidentification task (e.g., “총 3명 (a total of three people)” altered to “총 명수 1 (a total of one person)”), the token-level classification method of \sys eliminates such risks of sentence and context distortion (see Appendix ~\ref{app:llm}). Moreover, due to privacy and information security concerns, the use of API-based LLM services, such as ChatGPT, is restricted in many of the key government institutions in Korea ~\cite{NIS2023guidelines}. To create a trainable dataset from anonymized and unannotated court judgment data, we first manually label 6,700 civil, criminal, and administrative law cases that cover a broad spectrum of scenarios in civil, criminal, and administrative law. From these annotations, which identified 48,306 named entities, we establish a hierarchical categorization scheme for PII that aligns with relevant laws and practices and is suitable for model training. For each of the 729 labels in the PII scheme, we curate a corresponding list of entity mentions to generate model training data, as illustrated in \autoref{fig:overall}. Furthermore, we design a de-identification pipeline for the DNN-based language model, incorporating a specialized tokenizer that leverages the unique characteristics of the Korean language.

The approach used in this paper may offer valuable insights for other jurisdictions looking to efficiently anonymize large volumes of court decisions. The contributions of this paper are summarized as follows:
\begin{itemize}[nosep]
\item We have created a two-part dataset that consists of 6,700 labeled judgments from three kinds of cases: civil, criminal, and administrative cases and a list of actual entity mentions to replace the labels. The labeled judgments are created from publicly available anonymized court judgments.

\item We propose a three-tiered PII framework based on an inductive analysis of 48,306 named entities identified in our dataset.


\item  We propose a tokenizer that integrates a morphological analyzer, Mecab-ko, with Byte Pair Encoding (BPE) to leverage the unique features of the Korean language. Using this tokenizer, we also propose a method for generating training data from our labeled dataset and replacement list.

\item We evaluate \sys and it  achieves the highest performance among existing de-identification models for court judgments. 
\end{itemize}

	\section{Related Work}

Among others, there are many de-identification studies in health information. In the USA, de-identification in the medical field is guided by the Health Insurance Portability and Accountability Act (HIPAA)
~\cite{hhs_hipaa}, which defines two main strategies for compliance: the Safe Harbor method and Expert Determination ~\cite{meystre2010automatic, emelyanov2021towards}. The Safe Harbor method requires the removal of 18 identifiers called  Personal Health Information (PHI). Alternatively, Expert Determination relies on a statistical or scientific method to ensure minimal re-identification risk. In Europe, the General Data Protection Regulation (GDPR)~\cite{gdpr2016}
guides the de-identification of personal information in medical data. In this paper, we propose a three-tiered PII scheme for the de-identification of court judgment. 

\paragraph{Medical de-identification.} Research in medical de-identification has evolved through three major technical approaches. Early efforts primarily relied on rule-based systems ~\cite{uzuner2007evaluating}. With the advancement of deep learning, learning-based de-identification approaches, such as BiLSTM-CRF ~\cite{liu2017identification} and BERT-based NER models~\cite{berg2020impact, an2025identification}, were introduced. Large language models (LLMs) have been recently explored for de-identification in zero-shot or few-shot settings ~\cite{liu2023deid, altalla2025evaluating}. However, practical deployment is very limited because HIPAA regulations can be violated.

\paragraph{De-identification of court judgments.}
In recent years, there has been growing interest in automating the de-identification of court judgments based on NER. Many countries have launched government-led initiatives to adopt technical solutions to tackle problems with the labor-intensive de-identification procedure. The manual processing has been highlighted as delaying public disclosure and publication of judgments in Italy and Uruguay~\cite{salierno2024giusberto, garat2022automatic}. In India, the most populous country in the world, such a turn to automation is essential due to the overwhelming volume of court decisions~\cite{kalamkar2022named}. In Switzerland, automation has been introduced to assist court officials and legal experts in the anonymization process~\cite{niklaus2023automatic}. 
These NER-based methods report Precision, Recall, and F1-scores of 
96.43\%, 95.86\%, 96.14\% for Arabic~\cite{10283824}, 
92.26\%, 92.57\%, 92.40\% for German, French, and Italian texts (Switzerland), 
89.92\%, 90.50\%, 91.90\% for Spanish (Uruguay), 
92.00\%, 90.20\%, 91.10\% for Indian texts, 
and 85.00\%, 92.46\%, 88.60\% for Italian texts (Italy).

Having a substantial post-processing approach is critical in de-identifying court judgments. For instance, over-anonymization or unprincipled anonymization may undermine the readability of rulings when publicly disclosed~\cite{jpri2023openjustice}. The majority of previous studies~\cite{oksanen2022anonymization, niklaus2023automatic, salierno2024giusberto}  focus on how to detect personal identifiers in court judgments using NER, and less attention has been paid to discussing how the identified entities should be handled in the post-processing stage. Although the Uruguay study briefly addresses this issue, broader discussion and systematic approaches remain limited.

	\section{Methods}

\begin{table*}[!t]
\centering
\resizebox{0.55\linewidth}{!}{%
\begin{tabular}{llrr}
\hline
Domain & Case type & Documents & \textbf{Entities} \\
\hline
\multirow{4}{*}{Civil} 
 & 
 \begin{tabular}{l}
      Compensation for damage
 \end{tabular} & 901 & 9,223 \\
 &
   \begin{tabular}{l}
   Security deposit disputes 
     \end{tabular}  & 696 & 5,187 \\
 &  
  \begin{tabular}{l} 
   Payment of purchase price 
  \end{tabular} & 557 & 4,983 \\
 & \begin{tabular}{l} 
   Eviction 
  \end{tabular} & 846 & 6,816 \\
  \cline{2-4}
 & \begin{tabular}{l} 
   \textbf{Subtotal} 
   \end{tabular} & \textbf{3,000} & \textbf{26,209} \\
\hline
\multirow{5}{*}{Criminal} 
 & \begin{tabular}{l}
    Bodily injury 
      \end{tabular} & 600 & 2,562 \\
 & \begin{tabular}{l}
 Violence 
   \end{tabular} & 600 & 2,583 \\
 & \begin{tabular}{l}
 Sexual misconduct 
   \end{tabular} & 600 & 2,732 \\
 & 
 \begin{tabular}{l}
 Property theft \& deception 
   \end{tabular} & 600 & 4,376 \\
 & \begin{tabular}{l}
 Drunk driving 
   \end{tabular}& 600 & 2,354 \\\cline{2-4}
 & \begin{tabular}{l}
 \textbf{Subtotal } 
   \end{tabular} & \textbf{3,000} & \textbf{14,607} \\
\hline
Administrative 
& 
 \begin{tabular}{l} 
   Administrative litigation 
    \end{tabular}   
    & 700 & 7,490 \\ \cline{2-4}
& \begin{tabular}{l} 
  \textbf{Subtotal}
  \end{tabular}  & \textbf{700} & \textbf{7,490} \\
\hline
\textbf{Total} &  & \textbf{6,700} & \textbf{48,306} \\
\hline
\end{tabular}
}
\vspace{-0.5\baselineskip}
\caption{Number of documents and entities for each case type in the dataset.}
\label{tab:stats}
\end{table*}

There are three challenges unique to constructing datasets for the de-identification of court judgments in South Korea. First, since de-identification of court judgments prior to publication is a legal obligation of judicial institutions~\footnote{Korean Criminal Procedure Act, Art. 59-3; Korean Civil Procedure Act, Art. 163-2}, and only the fully anonymized judgments are available for external use, we need a method to generate datasets using anonymized and unannotated data. 
 
Second, there are legal rules to define categories of personal identifiers to be anonymized\footnote{Korean Supreme Court Regulation No. 2809 and Judicial Rule No. 1778}. However, they are not detailed enough to cover various attributes related to the persons involved in proceedings. They merely provide a direct identifier category and a broad quasi-identifier category that includes any other information that can identify the individual. 
 
Finally, since the South Korean judiciary heavily relies on manual de-identification, which is time-consuming ~\cite{congressional2019openpublication, nca2025courtpublication}, a large volume of court rulings that can immediately be used as a legal corpus for training is not available. 

\subsection{Data Collection}
\label{method:collection}

We initially compile 6,700 anonymized court decisions from a dataset provided by Korean Ministry of Government Legislation\footnote{https://www.moleg.go.kr/}, AI-hub\footnote{https://www.aihub.or.kr/} and \citet{hwang2022multi}\footnote{The dataset is released under the 
CC BY-NC 4.0 license.}. After removing duplicates across different sources, the final dataset comprises 3,000 civil, 3,000 criminal, and 700 administrative cases.

Our dataset encompasses a wide range of civil, criminal, and administrative scenarios, as summarized in Table~\ref{tab:stats}. By doing this, our dataset is better suited for identifying various types of domain-specific personal identifiers in court judgements.

We focus on collecting judgments rendered by courts of first instance. A significant portion of these judgments in Korea is dedicated to examining and clarifying facts, which is different from the approach taken in common law countries. At this level, the courts prioritize fact-finding and resolving disputed facts based on the investigations and evidence presented in court. 
Consequently, the collected judgments contain numerous direct and quasi-identifiers related to multiple individuals involved in the proceedings.

\subsection{Annotation Scheme}
We need a systematic annotation scheme for the annonymized court judgments to ensure that data labeling is consistent, reliable, and useful for our DNN-based de-identification process. The labeling process following the annotation scheme should be consistent across annotators and reproducible. The scheme should also speed up training for new annotators and helps maintain quality over large datasets. 

Without legal rules defining all relevant categories of personal identifiers, we develop an annotation scheme in four phases. First, human annotators identify placeholders (i.e., the anonymized sections in the judgment) in the provided text and label them using a set of entity categories we initially prepared based on an analysis of existing laws and practices. 
Second, while reviewing the labeling results for consistency among different annotators, we establish a new annotation scheme for PII with a \textit{three-tiered hierarchical structure} that classifies a range of entity types. Third, annotators make adjustments and corrections according to the annotation scheme. Finally, we resolve any issues where annotators may disagree or have doubts. 

\subsection{Placeholder Detection and Labeling} 
\label{method:labeling}
We have seventeen annotators who are fluent in Korean and possess a good understanding of NLP. They have completed an initial training session that provided guidelines on two main aspects: the key features of the task, which include a multi-stage process we designed for this project, and the rules regarding the scope and method of anonymization as applied in court practice.

Korean Judicial Rule No. 1778 establishes principles to guide court officials in using various de-identification methods. Depending on the type of identifiers involved, individuals can be represented with English letters (e.g., A and B) or combinations of letters (e.g., ABB, AAB). The complete removal of certain direct identifiers, such as resident registration numbers, is mandatory. For example, the text "... 피고인 홍길동 (561231-1234567) ... " ("... defendant Hong Gildong (561231-1234567) ...") would be anonymized to "... 피고인 A (주민등록번호 1) ..." ("... defendant A (resident registration number 1) ..."), where 561231-1234567 is a specific resident registration number. In this case, "피고인 A (주민등록번호 1)" represents the anonymized information obtained from the judgment within the collected corpus and is not labeled or annotated. The resident registration number is designated as 1 to differentiate between multiple individuals present in the judgment. 

Annotators manually identify the placeholders A and 1, labeling them to indicate the specific types of entities they represent as follows: "... $\lll$내국인이름$\ggg$A$\lll$/내국인이름$\ggg$($\lll$주민등록번호$\ggg$B$\lll$/주민등록번호$\ggg$) ...," where "내국인이름" refers to Korean names, and "주민등록번호" refers to a resident registration number. $\lll$내국인이름$\ggg$ and $\lll$/내국인이름$\ggg$ are markers and they point the beginning and end of the entity mention, respectively. "내국인이름" is a label to represent the category of the entity mention in our PII scheme (see Appendix ~\ref{app:sample}).

In an adjudication setting, locational information, such as the residential addresses of the parties involved in a case and the address of the crime scene, is essential for confirming the court's jurisdiction. It is standard practice to provide the exact address; however, under Korean Judicial Rule No. 1778, specific lower-level details of the address, like districts and streets, must be masked. At first glance, the address of a location or the name of a place may not seem like identifying information. However, their direct association with specific criminal activities can help identify the individuals involved in the case. Therefore, in accordance with existing laws and practices, lower-level address components and the names of all incident-related places must be de-identified. Similarly, contextual attributes such as the date of an event may also be considered quasi-identifiers and should be masked. For more examples of masking and labeling, please see Appendix~\ref{app:label}.

Annotators identified and labeled a total of 48,306 named entities across 6,700 court judgments.  \autoref{tab:stats} shows the number of documents and identified entities for the crime categories in the collected judgments. 

\subsection{PII Categories}
\label{method:PII}


As discussed earlier, existing law broadly defines the scope of de-identification. Aside from clear direct identifiers, quasi-identifiers often require more than just a textual assessment of the relevant attributes that can make an individual identifiable. The scope can be as extensive as "any other information identifying the persons involved in the case and third parties"\footnote{Korean Judicial Rule No. 1778, Art. 4}. Since it is nearly impossible to list all privacy-sensitive identifiers in writing, court officials are instructed to use discretion and analyze the specific context and its connection to the individuals involved in the case.

During the initial review of labeling, we found that many of the identified entities, specifically, the information anonymized in the collected judgments, do not consistently fit within the predefined categories of identifiers. There are challenging cases where the same type of entity may be evaluated differently across multiple judgments. 

For example, as a general rule, names of government institutions and public authorities (such as the Seoul Police Agency and the Seoul Correctional Institution) are not subject to de-identification. However, if these organizations are associated with the location where a crime was committed, exceptions may apply. This contextual interpretation of the case can lead to varying outcomes.

For another example, consider the following de-identified judgment text:"... 피고인 F와 피해자 G는 H 교도소 I 팀 소속의 교정공무원으로 ..." ("... defendant F and victim G were prison officers at team I of H correctional institution ...").  In actual de-identification practice, the name of the correctional institution ("교도소") is anonymized because it identifies the workplace where both the defendant and victim were colleagues. If this information is not anonymized in public disclosures, it could increase the chances of identifying the two individuals due to its direct connection to the circumstances surrounding the crime committed.

A different challenge in annotation arises when there are many individuals involved, and the specific roles each person plays in the case are not clearly defined during the anonymization process. This is particularly evident in cases of fraud, where a large group of victims is often targeted by illegal organizations, each member responsible for different aspects of the criminal activities. Additionally, in trials involving accomplices to a crime, it is crucial to anonymize identifiable information about various third parties, such as witnesses, appraisers, and forensic experts, to mitigate the risk of retaliation.

\begin{figure}[!t]
    \centering
    \includegraphics[width=\linewidth]{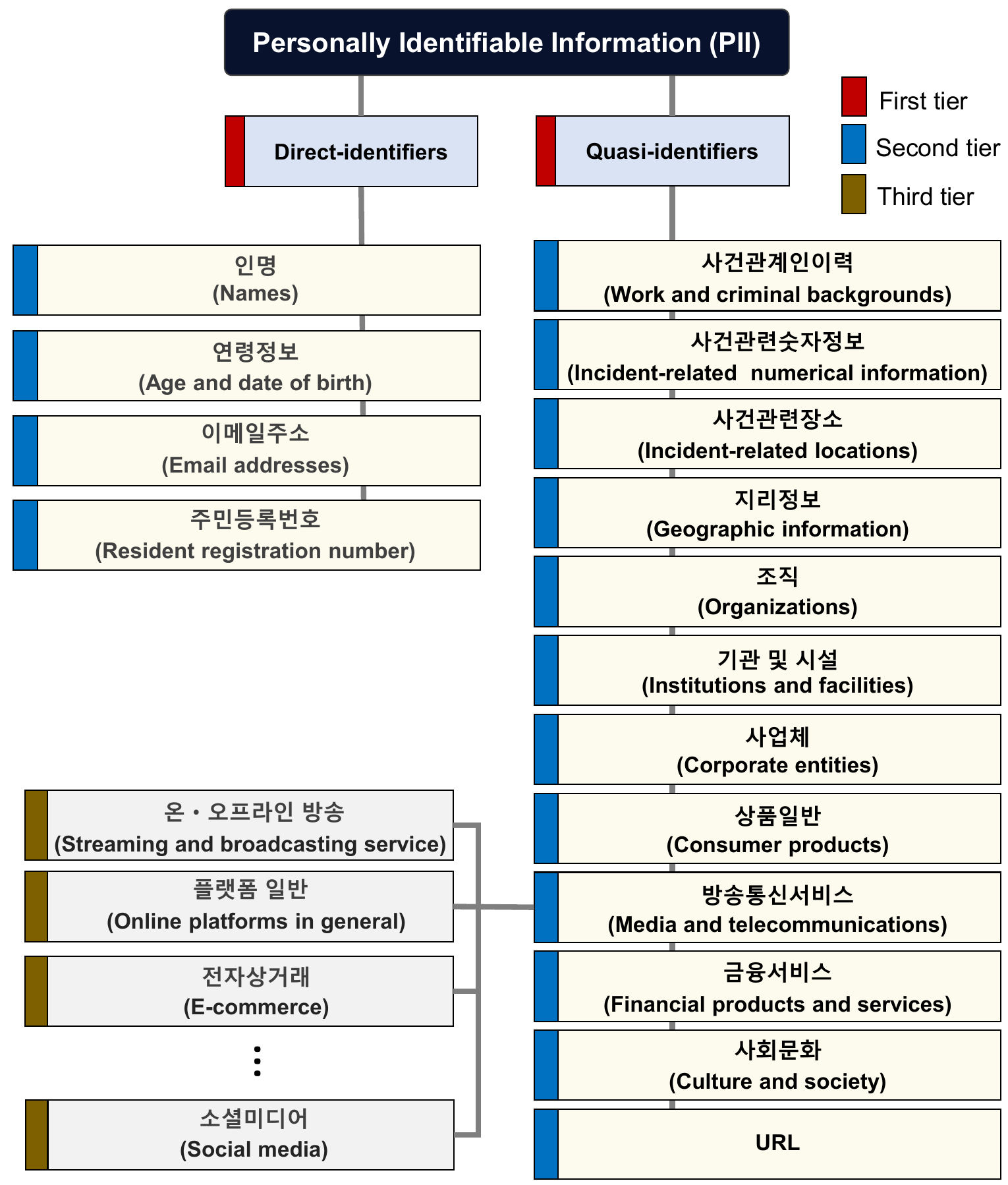}
      \vspace{-0.5\baselineskip}
    \caption{The three-tiered categorization scheme for PII in the domain of law and adjudication.}
    \label{fig:enter-label}
\vspace{-6mm}
\end{figure}

While the annotators made adjustments and corrections in accordance with the annotation scheme, we resolved any issues where the annotators disagreed or had uncertainties.

After reviewing all the named entities in the judgments, we developed our own PII annotation scheme that classifies various entity types into two main categories: direct identifiers and quasi-identifiers. This scheme includes 16 subcategories and 80 granular categories. Figure~\ref{fig:enter-label} illustrates the hierarchy of the categories. Each of the third-tier categories is associated with labels for annotation. Using this scheme, we annotated the identified named entities with a total of 729 labels. To the best of our knowledge, this is the first PII annotation scheme specifically designed for the de-identification of court judgments in Korea. Further details on the annotation scheme and its categories are provided in Appendix~\ref{app:scheme}. 

\subsection{Replacement Lists}
To improve the size and diversity of our training data, we create an extensive list of entity mentions using two different methods: manual curation and rule-based generation.

\paragraph{Manual curation.} We selectively choose reliable and verified information (entity mentions) sourced from the Korean government's licensing databases\footnote{\url{https://www.localdata.go.kr/main.do}} and public data portals\footnote{\url{https://www.data.go.kr/}}. We generate entity mentions for the majority of labels—691 out of 729. 
Our goal is to compile an average of at least 100 items for each label.

We also conduct searches on domain-specific websites to collect entity names related to specialized locations. For example, we gather lists of exhibition halls and conventions from the Coex Center, obtain names of ships and vessels from the Korea Seafarer's Welfare \& Employment Center, and collect names of junk dealers and recycling companies from the Korea Waste Recycling Institute. Additionally, we perform general web searches to supplement these results, ensuring a broad and diverse set of entity mentions that accurately reflect real-world usage.

\paragraph{Rule-based generation.} 
The second strategy uses rule-based generation to create entity mentions involving personal identifiers in standardized and structured formats. Simple rules are employed to generate entities such as Korean names, addresses in Korea, and numerical identifiers, which include resident registration numbers, phone numbers, and bank account numbers. 

\begin{figure*}[!t]
    \centering
    \includegraphics[width=0.95\textwidth]{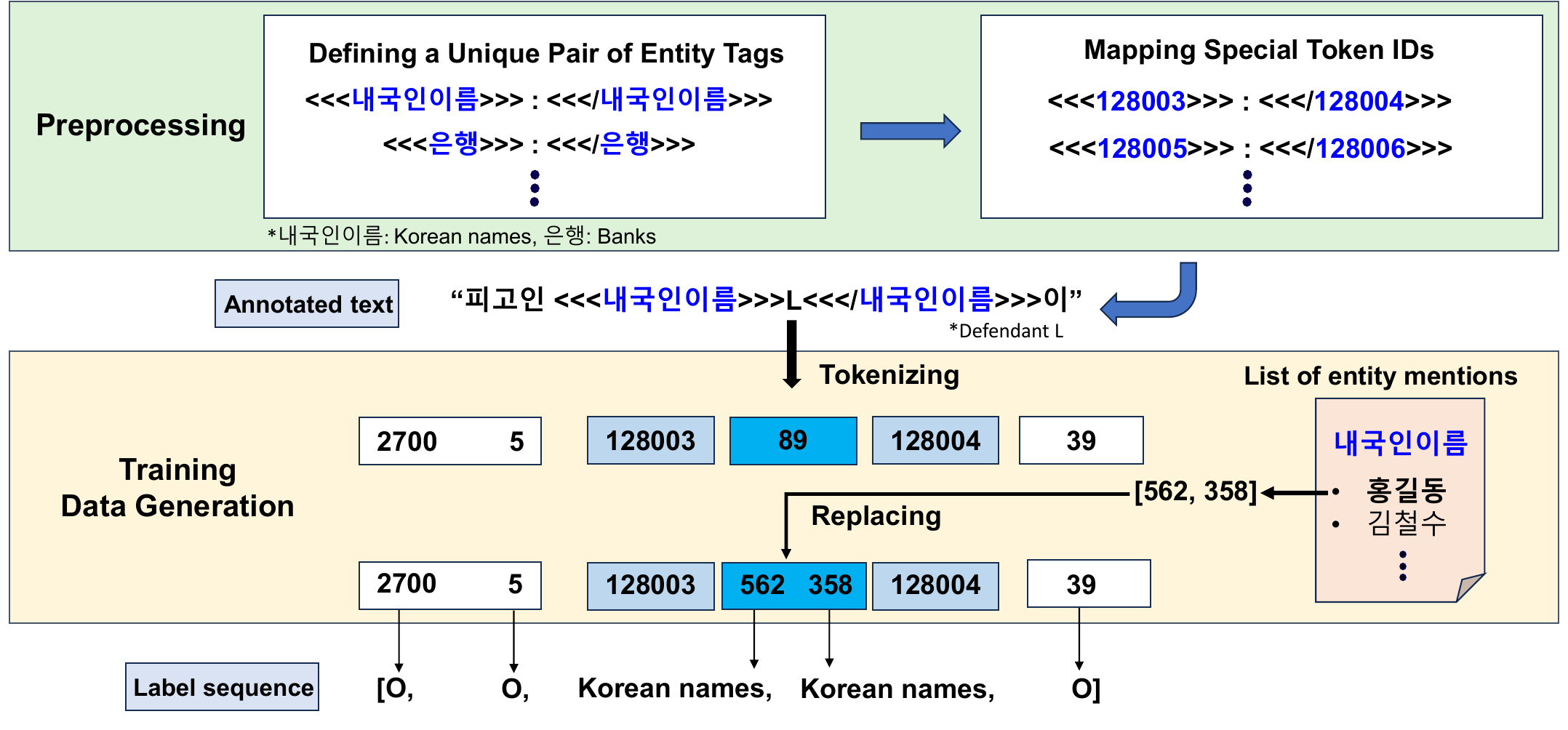}
    \vspace{-0.5\baselineskip}
    \caption{Tokenization and training data generation.} 
    \label{fig:tokenizer algorithm}
\vspace{-6mm}
\end{figure*}

\subsection{Training Data Generation} 
\label{sec:train_gen}
When we train our language model, we generate training data from the annotated dataset. In this process, we replace the labels in the dataset with actual entity mentions in the replacement list. Each labeled court judgment is augmented multiple times (N times) through entity mention replacements to maximize the amount of training data. 

For model training, documents are converted into tokenized input sequences (referred to as $X$) and corresponding label sequences (referred to as $Y$). Our tokenizer has been extended to include 1,458 special tokens that represent 729 different entities (labels). This extension is prioritized to ensure that proper nouns do not merge with other particles. Each judgment document is transformed into a token sequence, where subsequences marked with the start token $\lll$, placeholder tokens, and the end token $\ggg$ (e.g., "$\lll$name$\ggg$A$\lll$/name$\ggg$" after tokenization) are replaced with actual entity mention token sequences ("홍길동" after tokenization). Tokens within these subsequences are assigned the relevant label in $Y$ (e.g., "name") for de-identification, while tokens in other subsequences receive an "O" (outside) label in $Y$ to indicate that they do not require de-identification.

\subsection{Tokenization}  
We develop a custom tokenizer trained on a subset of one million sentences sampled from our corpus to effectively segment sensitive entities, such as names and organizations. Our tokenizer integrates a dictionary-based morphological analyzer, Mecab-ko\footnote{https://github.com/hephaex/mecab-ko}, with Byte Pair Encoding (BPE)~\cite{sennrich-etal-2016-neural}.

We choose Mecab-ko due to its ability to handle the Korean language's agglutinative morphology. It segments text into morphemes using a predefined dictionary, accurately distinguishing between nouns, particles, affixes, and adjectives. Studies have demonstrated Mecab-ko's effectiveness for recognizing domain-specific terms and proper nouns in Korean NLP tasks~\cite{Park2020, Cho2021, jeon2023improving}.

Unlike English, where proper nouns like "홍길동" remain unsegmented, Korean attaches nominative particles, such as "-이" and "-을," to nouns (e.g., "홍길동이"). Mecab-ko’s dictionary-based segmentation separates "홍길동이" into "홍길동" and "-이", ensuring that only the target entity ("홍길동") is de-identified while the particles remain intact. This approach helps the de-identified text flow smoothly and naturally. In addition, such precision is essential, given that the original (i.e., unanonymized and unannotated) court decisions lack clear boundaries for all entities.

While using a morphological analyzer like Mecab-ko is powerful, its fixed dictionary may not be able to capture rare legal terms or proper nouns, leading to out-of-vocabulary (OOV) issues. To overcome this limitation, we chose BPE, which builds a vocabulary through frequent character pair merges and represents unseen terms as subword units. 

\paragraph{Tokenization algorithm.}
The tokenizer recognizes special tokens and assigns unique token IDs to the beginning and end marker tokens of an entity mention. For instance, consider \autoref{fig:tokenizer algorithm}. We assign 128003 to \texttt{$\lll$}내국인이름$\ggg$ and 128004 to \texttt{$\lll$}/내국인이름$\ggg$. Given an input text from the annotated dataset, such as "피고인 \texttt{$\lll$}내국인이름$\ggg$L\texttt{$\lll$}/내국인이름$\ggg$이...", the text is tokenized into a sequence: [2700, 5, \underline{128003}, \underline{82}, \underline{128004}, 39], where L (token ID 82) serves as a placeholder for a labeled entity. Here, 내국인이름 refers to Korean names, and "피고인 L이" refers to "Defendant L".


Next, the token sequence is scanned to identify start marker tokens (e.g., 128003) and their corresponding end marker tokens (e.g., 128004), thus detecting the range of tokens between them. This range includes the placeholder (e.g., [128003, 82, 128004]). The placeholder within this range is then replaced with one of the entity mentions selected from the pre-defined replacement list. For example, in the sequence [2700, 5, 128003, 82, 128004, 39], the segment [128003, 82, 128004] is replaced by a token sequence [562, 358], which represents a name "홍길동" in the replacement list. This results in the updated sequence: [2700, 5, 562, 358, 39].

Subsequently, a corresponding label sequence is generated based on the indices of the replaced tokens, ensuring that the position and type of the labeled entity are retained (i.e., marking "홍길동" as a Korean names). For instance, the token sequence [2700, 5, 562, 358, 39] generates the label sequence [O, O, \underline{Korean names}, \underline{Korean names}, O], where "O" represents "Outside". This label sequence serves as the ground truth for supervised learning. Finally, the modified token sequence and its associated label sequence form a training data instance in the dataset (Figure~\ref{fig:tokenizer algorithm}).

\subsection{Data Augmentation} 

Due to the limited availability of publicly accessible court judgments, there will inevitably be instances where new entity types arise that the existing PII labels cannot represent. To address this limitation, we prepare a set of additional labels using LLM-assisted augmentation.

We begin by selecting specific granular categories that have significantly fewer labels compared to others. Next, we employ a large language model (LLM), such as ChatGPT~\cite{openai2022chatgpt}, to generate additional labels and create corresponding lists of entity mentions. For instance, "socio-cultural event" is one of the granular categories under  "Culture and Society" in the proposed PII scheme (Figure~\ref{fig:enter-label}). If, during the annotation process, we identify only a few labels within the "socio-cultural event" granular category, we can instruct the LLM to generate more labels for this category. Subsequently, we manually create several entity mentions for each additional label generated by the LLM.




\newcolumntype{C}{>{\centering\arraybackslash}p{1.5cm}}

\begin{table*}[t]
\small
\centering

\begin{subtable}{\linewidth}
\centering
\resizebox{0.95\linewidth}{!}{%
\begin{tabular}{l c | CCC | CCC}
\toprule

\multirow{2}{*}{\textbf{Model}} & \multirow{2}{*}{\textbf{\#Params}}
& \multicolumn{3}{c|}{\makecell[c]{\textbf{Single Replacement}\\\textbf{(Binary Token-Level)}}} 
& \multicolumn{3}{c}{\makecell[c]{\textbf{Per-Epoch Replacement}\\\textbf{(Binary Token-Level)}}} \\
\cmidrule(lr){3-5} \cmidrule(lr){6-8}
& & \textbf{Precision} & \textbf{Recall} & \textbf{F1} & \textbf{Precision} & \textbf{Recall} & \textbf{Micro F1} \\
\midrule
Polyglot-ko   & 1.3B & 0.9774 & 0.9570 & 0.9669 & 0.9710 & 0.9695 & 0.9701  \\ 
Exaone        & 2.4B & 0.9774 & 0.9542 & 0.9656 & 0.9688 & 0.9666 & 0.9677 \\
\midrule
\sys-360M     & 360M  & 0.9767 & 0.9264 & 0.9509 & 0.9628 & 0.9679 & 0.9654 \\
\sys-800M     & 800M  & 0.9786 & 0.9767 & 0.9776 & 0.9757 & 0.9826 & 0.9791 \\
\sys-1.5B     & 1.5B  & 0.9855 & 0.9683 & 0.9769 & 0.9755 & 0.9862 & \textbf{0.9808}  \\
\bottomrule
\end{tabular}
}
\subcaption{\textbf{Binary token-level} (Precision, Recall, and F1)}
\end{subtable}
\vspace{2mm}

\begin{subtable}{\linewidth}
\centering
\resizebox{0.95\linewidth}{!}{%
\begin{tabular}{l c | CCC | CCC}
\toprule
\multirow{2}{*}{\textbf{Model}} & \multirow{2}{*}{\textbf{\#Params}}
& \multicolumn{3}{c|}{\makecell[c]{\textbf{Single Replacement}\\\textbf{(Token-Level)}}}
& \multicolumn{3}{c}{\makecell[c]{\textbf{Per-Epoch Replacement}\\\textbf{(Token-Level)}}} \\
\cmidrule(lr){3-5} \cmidrule(lr){6-8}
& & \textbf{Precision} & \textbf{Recall} & \textbf{F1} & \textbf{Precision} & \textbf{Recall} & \textbf{Micro F1} \\
\midrule
Polyglot-ko   & 1.3B  & 0.8816 & 0.8631 & 0.8723 & 0.8772 & 0.8758 & 0.8765 \\ 
Exaone        & 2.4B  & 0.8785 & 0.8576 & 0.8679 & 0.8762 & 0.8742 & 0.8752  \\
\midrule
\sys-360M     & 360M  & 0.8895 & 0.8438 & 0.8660 & 0.8848 & 0.8895 & 0.8871 \\
\sys-800M     & 800M  & 0.9099 & 0.9082 & 0.9090 & 0.9073 & 0.9137 & \textbf{0.9105} \\
\sys-1.5B     & 1.5B  & 0.9091 & 0.8933 & 0.9011 & 0.9021 & 0.9120 & 0.9071 \\
\bottomrule
\end{tabular}
}
\subcaption{\textbf{Token-level} (Precision, Recall, and Micro F1)}
\end{subtable}
\vspace{-4mm}

\caption{
Performance comparison under different data generation settings.
Each sub-table reports \textbf{Precision, Recall, and F1} on the test set for the indicated evaluation granularity (Binary token-level vs Token-level).
Values are averaged over three random seeds (1200, 1203, 1205).
The best performance results are highlighted in bold.
}
\label{tab:perf-subtables}
\end{table*}

\section{Experiments}
\label{sec:experiments}
This section evaluates \sys and the experimental methodology.


\subsection{Experimental Setup}
\paragraph{Training datasets.} 
Besides our annotation dataset, we collect a bilingual corpus of approximately 76.7GB, comprising Korean and English texts from publicly available Web sources. This corpus is used for tokenizer training and pre-training for our language model. We also generate a dataset for NER-based de-identification using the method described in \autoref{sec:train_gen}. The dataset is divided into 80\% training (2,400, 2,401, and 560 documents), 10\% validation (300, 298, and 70 documents), and 10\% test (300, 301, and 70 documents), for civil, criminal, and administrative cases, respectively.

\vspace{-2mm}
\paragraph{Language models used.}
We train DeBERTa-v3-based models~\cite{debertav3}, \sys, with 370M, 800M, and 1.5B parameters for the de-identification of Korean court judgments through token classification. These models are compared against Korean-specialized language model baselines, namely Polyglot-Ko~\cite{polyglotko} and EXAONE-3.5~\cite{an2024exaone}, to assess their performance on our proposed dataset. For detailed information on the architectures and training configurations, please refer to Table~\ref{tab:model_config} in Appendix~\ref{app:training}.

\paragraph{Pre-training the models.}
\sys models are pre-trained from scratch using subsets of our bilingual corpus, which includes both English and Korean, containing 60 billion tokens for the 1.5 billion parameter model, 30 billion tokens for the 800 million parameter model, and 14 billion tokens for the 370 million parameter model. Training begins with a sequence length of 512 tokens, which is later extended to 2048 tokens to accommodate longer contexts. Unlike the original DeBERTa-v3, which uses post-LayerNorm, we adopt pre-LayerNorm~\cite{preln} because post-LayerNorm failed to converge for larger models, whereas pre-LayerNorm converged reliably under the same settings. For more details, please see Table~\ref{tab:model_config} in Appendix~\ref{app:training}.

\vspace{-3mm}
\paragraph{Fine-tuning the models.}
\sys models and the baseline models were fine-tuned on our dataset, which consists of 5,361 training documents (2,400, 2,401 and 560), 668 validation documents (300, 298, 70) and 671 test documents (300, 301, 70) for civil, criminal, and administrative cases, respectively. We employ both Per-Epoch and Single Entity Replacement methods to assess the effects of data variation. The training use a sequence length of 2,048 tokens over the course of 30 epochs. For detailed training information, please refer to Table~\ref{tab:model_config} in Appendix~\ref{app:training}, and for the results, see Table~\ref{tab:perf-subtables}. 

\paragraph{Evaluation metrics.}
We use three metrics — precision, recall, and F1-score — to assess the performance of our model on the de-identification task. Each metric is evaluated under two settings: binary token-level~\cite{dernoncourt2016deid, yuezhou2020phicon, salierno2024giusberto, kim2024generalizing} and token-level~\cite{dernoncourt2016deid, yuezhou2020phicon, kim2024generalizing}. The binary token-level setting measures the model’s ability to correctly classify tokens that require de-identification and those that do not, without considering the type of entity. For the details of the two settings and metric definitions, please see \autoref{app:metrics}.
\vspace{-2mm}
\subsection{Experimental result}
\label{sec:exp-results}
Table~\ref{tab:perf-subtables} shows the performance of our models compared to two Korean-specialized Decoder models, Polyglot-ko (1.3B) and Exaone (2.4B), under two data generation settings: Single Replacement and Per-Epoch Entity Replacement. \sys models consistently outperform the baselines in both binary token-level and token-level micro F1 scores. Our largest model \sys-1.5B achieves a binary token-level F1 of 0.9808 and 800M model achieves a token-level F1 of 0.9105 under the Per-Epoch Entity Replacement setting, establishing a state-of-the-art (SOTA) benchmark for NER-based de-identification of Korean court judgments. Notably, even our smallest model \sys-370M (0.8871) outperforms both Polyglot-ko (0.8765) and Exaone (0.8752) in the token-level micro F1 metric. For a detailed breakdown of performance by case type, please refer to Appendix~\ref{app:case-results}.

The high binary token-level F1 score for \sys under Per-Epoch Entity Replacement demonstrates that the model is proficient in identifying which tokens need to be de-identified. Additionally, the high token-level micro F1 score indicates that \sys effectively classifies the entity types of these de-identifiable tokens. 
Given that the model is required to classify as many as 729 distinct labels, achieving a token-level F1 score exceeding 0.91 is a strong indicator of its robust multi-class classification performance.

The Per-Epoch Entity Replacement technique significantly outperforms Single Replacement in all models, including Polyglot-ko and Exaone. This consistent improvement highlights the quality of our dataset, its annotation scheme, and the corresponding list of entity mentions for realistic value generation. Frequent entity replacements enhance data diversity while maintaining high-quality augmentation and effective generalization.


The 800M model demonstrates a slightly higher token-level micro F1 score under the per-epoch setting compared to the 1.5B model. In our data-limited scenario, the 800M model may be better suited to the dataset size, allowing it to generalize slightly better. In contrast, the 1.5B model may overfit to rare labels. However, the difference between the two models is minimal and could diminish with additional data for rare labels or the application of stronger regularization.

\sys demonstrates weaknesses in identifying low-frequency labels that seldom appear in the training corpus. For example, it frequently misclassifies “뷔페 (buffet restaurant)”—which should fall under “외식업 (eating and drinking places)”—as “기계설비회사 (machinery and equipment company)” within the “제조업 (manufacturing)” category. As our annotators reviewed the fully anonymized court judgments, we noted some exceptional cases where it was challenging to accurately determine the exact type of entity, despite careful contextual analysis. These instances also resulted in misclassifications, such as labeling “불특정제품명 (unspecified product name)” under “상품 일반 (general products)” as “불특정회사명 (unspecified company)” under “기업 일반 (companies and businesses in general).”

\sys significantly outperforms the rule-based system currently used by the Korean National Court Administration, which reportedly achieves an overall accuracy of 8 to 15\% ~\cite{congressional2019openpublication, nca2025courtpublication}. These results position \sys as a new and effective framework for Named Entity Recognition (NER)-based de-identification of court judgments.






\section{Conclusion}
\vspace{-2mm}
In this paper, we propose a DNN-based solution, referred to as \sys, for NER aimed at improving the efficiency and consistency of de-identifying court judgments. We address the complex challenges currently faced in the de-identification process within the Korean judiciary. Our work includes the development of the first Korean legal dataset, which contains 6,700 judgments from civil, criminal, and administrative cases, encompassing a total of 48,306 labeled named entities. We also introduce a three-tiered annotation scheme for PII, which systematically categorizes a wide variety of personal identifiers. Furthermore, we provide a comprehensive list of entity mentions that can be used to replace the 729 token-level labels found in the training dataset. In addition, we outline a tokenization method for the training data generated from these replacements. Our experimental results show that \sys achieves state-of-the-art performance in the de-identification of court judgments.

    \clearpage
\section*{Limitations}
Our study has some limitations. First, original (unanonymized) court judgments are not accessible due to legal restrictions. As mentioned earlier, we only have access to fully anonymized judgments that have been processed and reviewed by court officials before being made public. This limitation prevents us from evaluating our model's performance in real-world settings. To address this issue and make our model more applicable to actual de-identification practices within the Korean judiciary, we plan to develop a more strategic method of data augmentation for future research. This includes creating synthetic data that closely resembles court judgments. By pursuing this direction, we aim to increase the size and diversity of our training data, allowing for more robust testing of our model. 

Second, our model was specifically trained using judgments from the field of civil, criminal, and administrative law and procedure. De-identification in the legal domain is highly context-sensitive, which means the model's performance may decrease when applied to court decisions involving different types of legal disputes. 
However, we anticipate that our model will still perform reasonably well, as there are shared characteristics regarding direct identifiers across various types of court judgments. Additionally, our dataset encompasses a wide range of entity types. Thus, our system has important implications even for court judgments in entirely different areas of law. Further research is necessary to evaluate the model's performance in these other areas and to explore how the proposed method can be adapted and enhanced for effective de-identification tasks across diverse legal contexts.

    \section*{Ethics Statement}


All court judgments used in this study were obtained from publicly available anonymized datasets, including those released by the Korean Ministry of Government, AI-hub\footnote{https://www.aihub.or.kr/} and published by \citet{hwang2022multi}, none of which contain any PII. To support data reconstruction and model training, replacement lists were compiled exclusively from open-access sources, including government licensing databases\footnote{\url{https://www.localdata.go.kr/main.do}}, public data portals\footnote{\url{https://www.data.go.kr/}}, and official institutional websites\footnote{\url{https://data.seoul.go.kr/}}. No private or sensitive information was used at any stage of this research.

Although the dataset is fully anonymized and all sources are publicly available, we ensured that our data processing procedures—including the creation of replacement lists—adhered to the principles of the Korean Personal Information Protection Act (PIPA).

    \section*{Acknowledgements}


We thank the anonymous reviewers and the meta-reviewer for their valuable feedback on this paper.
We also sincerely thank Gyeongje Cho, Hyeonggeun Jeon, Sungmok Jung, Dayeon Kang, Jia Kang, Minsu Kim, Sangho Kim, Jongmin Kim, Dongyoung Lee, Joonhak Lee, Changjin Lee, Jongyeon Park, Yoonhee Park, Seho Pyo, Jiheon Seok, Yeonkyoung So, and Youngjun Son for their dedicated work as annotators.

This work was partially supported by the National Research Foundation of Korea (NRF) under Grant No. RS-2023-00222663 (Center for Optimizing Hyperscale AI Models and Platforms), and by the Institute for Information and Communications Technology Promotion (IITP) under Grant No. 2018-0-00581 (CUDA Programming Environment for FPGA Clusters) and No. RS-2025-02304554 (Efficient and Scalable Framework for AI Heterogeneous Cluster Systems), all funded by the Ministry of Science and ICT (MSIT) of Korea. It was also partially supported by the Korea Health Industry Development Institute (KHIDI) under Grant No. RS-2025-25454559 (Frailty Risk Assessment and Intervention Leveraging Multimodal Intelligence for Networked Deployment in Community Care), funded by the Ministry of Health and Welfare (MOHW) of Korea. Additional support was provided by the BK21 Plus Program for Innovative Data Science Talent Education (Department of Data Science, Seoul National University, No. 5199990914569) and the BK21 FOUR Program for Intelligent Computing (Department of Computer Science and Engineering, Seoul National University, No. 4199990214639), both funded by the Ministry of Education (MOE) of Korea. This work was also partially supported by the Artificial Intelligence Industrial Convergence Cluster Development Project, funded by the MSIT and Gwangju Metropolitan City. Research facilities were provided by the Institute of Computer Technology (ICT) at Seoul National University.

￼
￼
￼

	
	
\bibliography{custom_arxiv}

 \clearpage
\appendix
\appendix 

\setcounter{figure}{0}  
\setcounter{table}{0} 
\counterwithin{figure}{section}
\counterwithin{table}{section}

\section*{\centering\textbf{Appendix}}

\section{Issues in Prompt-based De-identification}
\label{app:llm}
We identify the following five categories of problems frequently appearing in the GPT-assisted de-identification dicussed in Section~\ref{sec:intro}. These cases represent the ways in which prompting-based anonymization can lead to compromise textual integrity of public records and undermine legal precision required for settling disputes effectively.
\begin{itemize}
    \item First, rewriting and paraphrasing frequently occurred. For example, the verb “입금하였다 (deposited)” was changed to “송금하였다 (wire transferred).” While both can describe sending money to someone, the forms and implications of these behaviors are differently conceived in legal and financial contexts.
    \item Second, we also found cases of partial omission when GPT removed, for instance, the phrase “제때 (on time)” from the original text. The original phrase “그 대금을 제때 변제하여” (“by repaying the amount on time”) was shortened to “대금을 변제하여” (“by repaying the amount”) in GPT-4’s output. The omission of “제때” (“on time”) removes an important indication of timely payment, which is often critical in determining whether the legal obligation was properly met.
    \item Third, (unsolicited) summarization of the original text resulted in the loss of detailed facts and strategies concerning the crimes committed. Unlike the original text, it merely provides a brief summary of the factual backgrounds of the case. For instance, after going through GPT-assisted de-identification, three sentences containing important details about defendant’s intention and plan to defraud victim and the amount of damage caused were vaguely summarized and reduced to a single sentence, “피고인은 이를 개인 용도로 사용하였다 (The defendant used it for personal purposes)”.
    \item Fourth, in the cases where multiple individuals and institutions are involved in the litigation, we often identified entity collapse: a number of different entities were anonymized with the same letter (e.g., 광주은행 (Gwangju Bank), 우정사업본부 (Korea Post), 부산은행 (Busan Bank) → A, A, A).
    \item Lastly, distortion of facts occurred. For example, specific numbers in the judgment were altered during de-identification “총 3명 (a total of three people)” was altered to “총 명수1 (a total of one person)”.
    
\end{itemize}

Moreover, due to privacy and information security concerns, the use of API-based LLM services such as ChatGPT is restricted in Korean government institutions. Domestic regulations (issued by the National Intelligence Service and the Ministry of the Interior and Safety) require public officials across government departments to refrain from putting in any sensitive internal data and personal information while using such services.
\section{Data Samples}
\label{app:sample}
Since it is a legal obligation of the courts to anonymize judgments prior to public disclosures, there is no way to access unannoymized judgments which could have served as ground truth for our research. After collecting fully anonymized judgments, we manually annotate the whole corpus based on the three-tiered categorization scheme classifying a range of personal identifiers.  (See Section ~\ref{method:labeling})

\begin{figure*}
    \centering
    \includegraphics[height=0.95\textheight, keepaspectratio]{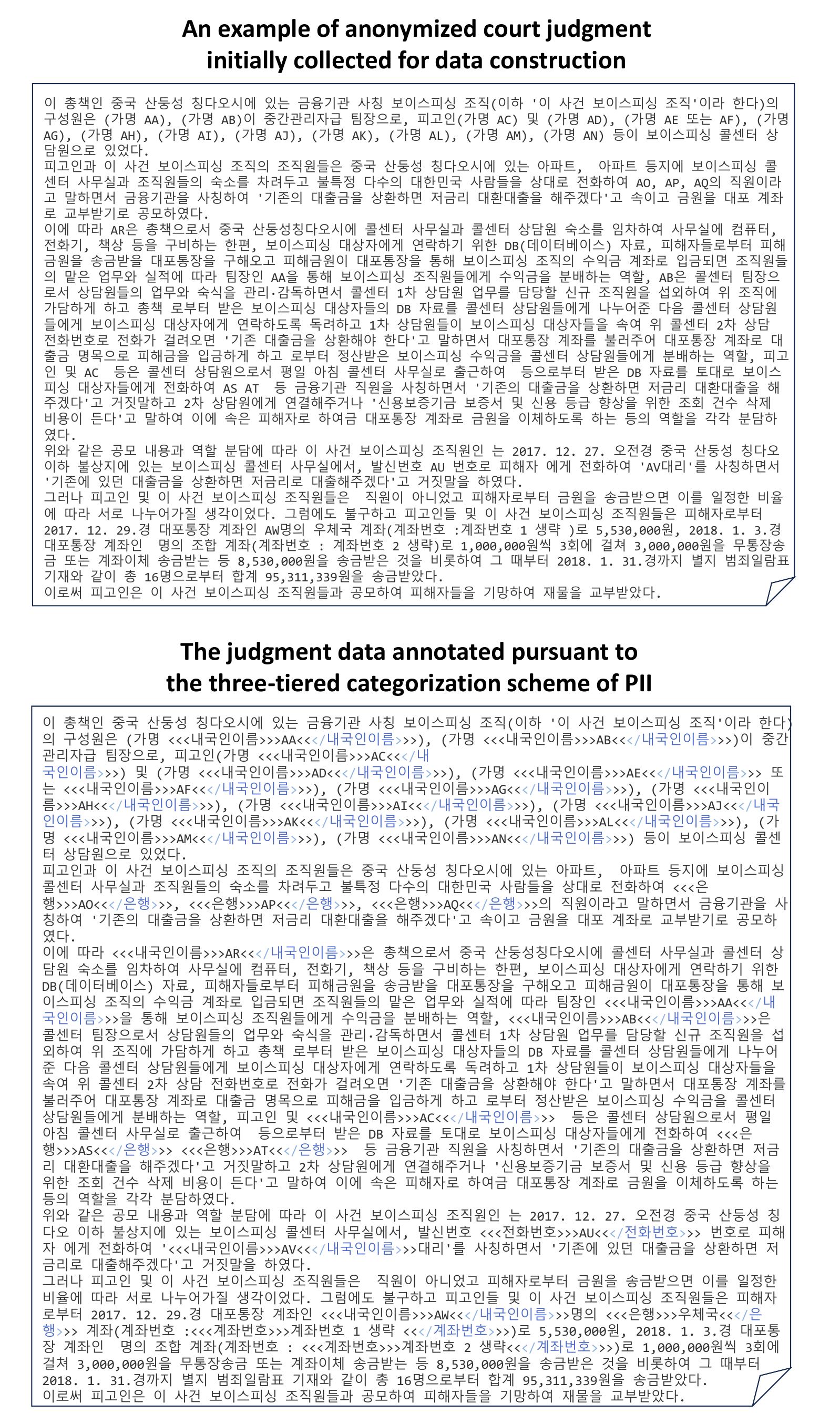}
    \caption{Examples of court judgment data before and after annotation.}
    \label{fig:example}
\end{figure*}

To give our readers the gist of the collection and annotation process, Appendix C presents one of the examples of the court judgment initially compiled for data construction as in \autoref{fig:example}.  Next to this anonymized court judgment, the annotated version of that same judgment appears.





\section{Masking and Labeling Examples}
\label{app:label}
Appendix B illustrates the example discussed in Section~\ref{method:labeling} in more detail.

\paragraph{Example with a functional descriptor.}
Locational information in the sentence “… 나리 식당에서 근무하는 피고인 홍길동은 …” can be de-identified as “… A 식당에 근무하는 피고인 B는 …”. According to Korean Judicial Rule No. 1778, “나리 식당” qualifies as identifying information due to its contextual specificity. While the generic term “식당” (diner) remains intact as a non-identifying functional descriptor, the unique component “나리” is replaced with the placeholder “A”. Similarly, the name “홍길동” is replaced with “B”. The resulting sentence is labeled as:

“… $\lll$식당$\ggg$A$\lll$/식당$\ggg$ 식당에 근무하는 피고인 $\lll$내국인이름$\ggg$B$\lll$/내국인이름$\ggg$는 …”,
where the label \texttt{식당} refers to a place for eating and drinking.

\paragraph{Example without a functional descriptor.}
In constrast, some place names do not contain an explicit functional descriptor. For example, the sentence "... 피해자 김철수를 기다리며 맥도날드에서 음식을 주문하고 ..." ("... ordered food at McDonald's while waiting for the victim Kim Chulsoo to arrive ...") can be anonymized to "... 피해자 D를 기다리며 E에서 음식을 주문하고 ..." ("... ordered food at E while waiting for the victim D to arrive ..."). In this case, "맥도날드" ("McDonald's") does not have a functional descriptor, so court officials are instructed to replace the entire word with a placeholder, E. Therefore, "... 피해자 D를 기다리며 E에서 음식을 주문하고 ..." in the de-identified judgment will be labeled by the annotators as follows: "... 피해자 $\lll$내국인 이름$\ggg$D$\lll$/내국인 이름$\ggg$를 기다리며 $\lll$식당$\ggg$E$\lll$식당$\ggg$에서 음식을 주문하고 ...."

\section{Personally Identifiable Information (PII) Categorization}
\label{app:scheme}

Appendix B provides a complete overview of the three-tiered categorization scheme classifying personal identifiers in the domain of law and adjudication, as detailed in Section~\ref{method:PII}. Under two main categories, 16 subcategories, and 80 granular categories, we present a total of 729 labels alphabetically ordered in Korean along with the English translation of each label. 


\subsection{사건관계인 특정 정보 (Direct identifiers)}
\subsubsection{인명 (Names)}
\paragraph{내국인이름 (Korean names):}
\paragraph{외국인이름 (Non-Korean names):}
몽골인이름 (Mongolian names), 베트남이름 (Vietnamese names), 세례명 (baptismal names), 영어이름 (English names), 일본인이름 (Japanese names), 중국인이름 (Chinese names), 캄보디아이름 (Cambodian names), 태국인이름 (Thai names), 필리핀이름 (Filipino names), 러시아권이름(Russian names), 법명(Dharma names)
\\ 

\paragraph{아이디\textbullet 닉네임 (IDs and Nicknames):}
가수 (aliases), 닉네임 (nicknames), 대화명 (usernames), 별명 (nicknames), 블로그 (blogs), 아이디 (IDs), 법호(Dharma nickname)
\subsubsection{연령정보 (Age and Date of Birth)}
나이(age), 출생연도(year of birth), 생년월일(date of birth)
\subsubsection{이메일주소 (Email Address)}
이메일주소 (email address)
\subsubsection{주민등록번호 (Resident Registration Number)}

\subsection{기타 (사건관계인이나 제3자를 특정할 수 있는) 정보 (Quasi-identifiers)}

\subsubsection{사건관계인이력 (Work and Criminal backgrounds of the persons involved in the case)}
\paragraph{범죄경력 (Criminal records):}
죄 (crime)
\subsubsection{사건 관련 숫자 정보 (Incident-related numerical information)}
\paragraph{고유번호 (Various Numbers Uniquely Identifying Specific Individuals and Objects):}
계좌번호 (bank account number), 관리번호 (management number), 금괴일련번호 (gold bar serial number), 사건번호 (case number), 선박번호 (IMO ship number), 비트코인개인지갑 (bitcoin wallet), 수표번호 (check number), 카드번호 (card number), 어선 (fishing vessel number), 어음번호 (bill number), 범죄경력등조회회보서 (criminal record certificate), 차량번호 (vehicle registration number), 특허번호 (patent number), 휴대폰번호 (mobile phone number), 군번 (military service number), 면허번호 (license number), 훈장번호 (decoration number), 전화번호 (phone number), 내선번호 (extension number), 수험번호 (examination number), 보훈번호 (veterans registration number), 보증번호 (guarantee number), 고시번호 (official notice number), 비밀번호 (password / PIN), 등기번호 (registration number), 사업자등록번호 (business registration number), 접수번호 (receipt number), 민원번호 (civil complaint number), 경매번호 (auction number), 채권번호 (bond number), 일련번호 (serial number), 법인등록번호 (corporate registration number)

\paragraph{장소 관련 번호 (Numbers Assigned to  Specific Places):}
골프장코스 (golf course), 구역 (zone), 라인 (line), 지하철칸 (subway compartment), 항공편 (flight number)
번(number), 호선(line number), 호실(room number), 호(unit number), 출구번호(exit number), 동(building number), 층(floor), 노선번호(route number), 레일(rail number), 승강장번호(platform number), 열차번호(train number), 탑승장번호(boarding platform number), 번호(number), 광역버스(express bus number), 단지 (housing complex), 로트 (lot), 블록 (block), 실(room), 번지(lot address number)

\paragraph{기타 사건 관련 숫자 (Other Incident-related Numbers):}
기수 (class number), 명수(number of people), 연도(year), 날짜(date)

\subsubsection{사건 관련 장소 (Incident-related sites and locations)}
\paragraph{시설 내부 공간 (Interior Spaces):}
건물내장소 (a place in the building), 공공기관내장소 (a place in the public institution), 공원내장소 (a place in the park),   
광장(square), 소분류장(small classification yard), 사무실(office), 
교도소내장소 (a place in the correction facility), 구치소내장소 (a place in the detention center), 대학교내장소 (a place in the university), 문 (gate),물류센터레일 (rails at logistics center), 법원내장소 (a place in the courthouse), 병원내장소 (a place in the hospital), 생활관 (residential hall), 아파트내장소 (a place in the apartment), 기숙사 (dormitory), 군부대내장소 (a place in the military facility)
\paragraph{교통 (Transport Infrastructure):}
버스공영차고지 (bus garage)
버스정류장 (bus stop), 요금소 (tollgate)

\paragraph{건설 (Construction Sites):}
공사장 (construction yard), 현장 (site), 야적장 (storage yard)
공사현장 (construction site) 
\paragraph{산림\textbullet 하천 (Forest and Water):}
둘레길 (perimeter trail), 등산로 (hiking path), 산책로 (walking trail), 약수터 (mineral spring)
\paragraph{해양 (Places related to Maritime Activities):}
선박명 (ship name), 여객선 (passenger ship name), 군함명 (warship name)

\subsubsection{지리정보 (Geographic information)}
\paragraph{주소 (Address):}
도아래주소 (address under province)
구아래주소 (address under district/Gu)
군아래주소 (address under county/Gun)
읍아래주소 (address under town/Eup)
동아래주소 (address under neighborhood/Dong)
시아래주소 (address under city/Si)
주소 (address)
임야 (forest land)
토지 (land)
필지 (parcel/lot)
국외주소 (overseas address)

\paragraph{지역명 (Geographic units):}
마을 (village), 산 (mountain), 선거구 (coinstituency), 선거단위 (electoral district), 외국도시 (foreign city), 지구 (district), 해안지역명 (coastal area name), 
해수욕장 (bathing beach), 국외하천 (overseas river/stream), 
행정구 (Gu: district-level administrative unit), 
행정군 (Gun: county-level administrative unit), 
행정동 (Dong: neighborhood-level administrative unit), 
행정리 (Ri: village-level administrative unit), 
행정면 (Myeon: township-level administrative unit), 
행정시 (Si: city-level administrative unit), 
행정읍 (Eup: town-level administrative unit), 행정도 (Do: province-level administrative unit), 베트남전관련지명 (Vietnam War related place names), 지사및지청명 (branch and local office names), 
특정지역범위명 (specific regional boundary names), 고지 (highland/hill), 국가명 (country name), 고개 (mountain pass), 섬이름 (island name), 전투지역(combat zone), 해변 (beach), 호수(lake), 하천 (river/stream), 

\paragraph{도로명 (Roads and Streets):}
골목 (alley), 교차로 (intersection), 길 (street), 도로 (road), 인터체인지 (interchange), 로터리 (rotary)

\paragraph{구간 (Sections)}
도로구간 (road section), 철도구간 (railway section)

\subsubsection{조직 (Organizations)}
\paragraph{친목\textbullet 문화 (Community Gatherings):}
단체명 (uncategorized gatherings), 독서토론모임 (book club), 동호회 (uncategorized clubs), 모임 (social gatherings), 봉사단체 (volunteer group), 산악회 (hikers club), 연합회 (uncategorized coalitions), 체육회 (sports club)
\paragraph{사회\textbullet 종교 단체(Social and Religious Groups):}
노회 (presbytery), 사회복지법인 (social welfare organization), 종교단체 (religious organization), 종중 (class association)
\paragraph{정치\textbullet 경제 단체 및 협의체 (Various Associations of Like-minded People in Politics, Commerce and Labor):}
공제조합 (mutual aid association), 노동조합 (labor union), 선거캠프 (election camp),  재개발정비조합 (redevelopment partnership), 재건축정비조합 (reconstruction partnership),
정당 (political party), 조합 (uncategorized partnerships), 지역주택조합 (local housing association), 협동조합(cooperative association), 협의회 (uncategorized councils), 협회 (uncategorized associations), 상인회 (merchant association), 사단법인 (non-profit corporation), 의료법인 (medical corporation), 위원회 (committee), 재단법인 (foundation), 학교법인 (educational foundation), 의료재단 (medical foundation), 어촌계 (fishermen’s association), 총회 (general assembly)

\paragraph{국방\textbullet 치안 (Specific Units in Military and Law Enforcement Agencies):}
국정원비밀조직 (secret agency under the National Intelligence Service), 
대대 (battalion), 
헌병대 (military police), 사단 (division), 여단 (brigade), 소대 (platoon), 연대 (regiment), 중대 (company), 사령부 (headquarters), 해군전단 (naval squadron), 본부 (headquarters), 해군함대 (naval fleet)

\paragraph{조직 내 세부부서 (Specific Units and Departments in the Organizations):}
단과대학 (college), 반 (kindergarten class), 부서 (departments), 지회 (branches), 팀 (teams), 학과 (college majors), 교통공사내부서 (department within transportation corporation)

\paragraph{조직 내 업무\textbullet권한 등(Job levels and duties within organizations)}
직급 (job level), 회원등급 (membership level), 군계급 (military rank), 직무 (job duty), 보직 (official post)

\paragraph{불법 단체 (Illegal Organizations):}
범죄조직 (criminal organization)

\subsubsection{기관 및 시설 (Institutions and Facilities):}
\paragraph{정부기관 및 지방자치단체 (Public Administrative Bodies and Local Municipalities):}
공사및공단 (public institution), 시청 (city hall), 우체국 (post office), 행정복지센터 (community service center), 중앙행정기관 (central administrative agency), 해양및산림등관리기관 (maritime and forestry management agency), 교육청 (office of education), 등기소 (registry office), 세무서 (tax office)

\paragraph{군사 (Military Bases):}
군부대 (military camp), 미군부대 (US Army), 훈련소 (military training center), 군정비및관리시설 (military maintenance and management facility), 군소속교육기관 (military-affiliated educational institution)
 
\paragraph{치안 및 교정 (Policing and Correctional Facilities):}
경찰서 (police office), 경찰청 (national police agency), 구치소 (detention center), 지구대 (police substation), 치안센터 (community police center), 파출소 (police substation) 
\paragraph{소방 및 재난 (Agencies for Fire Safety and Disaster Response):}
소방서 (fire station), 안전센터 (safety center)
\paragraph{사회기반시설 (Public Infrastructure):}
공항 (airport), 발전소 (power plant), 버스터미널 (bus terminal), 선착장 (dock), 육교 (pedestrian overpass), 저수지 (resorvoir), 지하차도 (underpass), 지하철역 (subway station), 태양광발전소 (solar power plant), 터널 (tunnel), 항구 (port), 교량 (bridge), 비행장 (airfield), 검사및검문소 (inspection and checkpoint), 상하수도시설 (water supply and sewage facilities), 변전소 (substation), 원자력본부 (nuclear headquarters), 부두 (pier/wharf), 기차역 (train station)

\paragraph{사회복지시설 (Social Security and Welfare Facilities):}
복지시설 (welfare facility), 요양원 (nursing home), 육아원 (child care center), 장애인이용시설 (facility for person with disabilities), 재가장기요양기관 (home-based long-term care institution)
\paragraph{주민편의시설 (Residential Convenience Facilities):}
공원 (park), 농어촌근린시설 (rural community facility), 마을회관 (community center), 주민쉼터 (community rest area), 경로당 (senior center), 유원지 (recreational area), 놀이터 (playground), 마당 (yard)

\paragraph{스포츠시설 (Sports Facilities)}
경기장 (stadium), 야구장 (baseball stadium)

\paragraph{주거시설 (Residential Buildings):}
고급주택 (luxury residence), 맨션 (low-rise apartment), 빌라 (multiplex housing), 아파트 (apartment), 오피스텔 (studio apartment), 주택 (single-family home), 타운하우스 (townhouse)

\paragraph{의료기관 (Healthcare Institutions):}
내과 (internal medicine clinic), 병원 (hospitals), 산부인과의원 (OB-GYN clinic), 성형외과 (plastic surgery), 신경외과 (neurosurgery), 안과 (ophthalmalmic clinic), 요양병원 (nursing hospital), 의원 (local clinic), 정신병원 (mental hospital), 정형외과 (orthopedics clinic),
치과 (dentistry), 치과의원 (dental clinic), 의료원 (medical center), 보건소 (public health center), 재활원 (rehabilitation center), 이비인후과 (ENT clinic), 피부과 (dermatology clinic)
한방병원 (Korean medicine hospital), 한의원 (oriental medicine clinic)

\paragraph{교육기관 (Educational Institutions):}
고등학교 (highschool), 대학교 (university), 어린이집 (daycare center), 연수원 (training center), 유치원 (kindergarten), 중학교 (middle school), 직업능력개발훈련시설 (vocational training center), 초등학교 (elementary school), 국외중고등학교 (overseas middle and high school), 국외대학교 (overseas university), 사관학교 (military academy), 전문학교(vocational school), 중고등학교 (secondary school)

\paragraph{문화\textbullet 예술 (Art and Cultural Facilities):}
도서관 (library), 문화시설 (culture center), 미술관 (art museum), 전시장 (exhibition hall), 청소년수련관 (youth training center), 박물관(museum)
\paragraph{종교시설 (Place of Worship):}
교회 (church), 사찰 (temple) 
\paragraph{상업시설 (Commercial Buildings and Facilities):}
빌딩 (building), 상가 (shopping plaza), 시장 (market), 아울렛 (outlet),
장례식장 (funeral home), 
전기차충전소 (EV charging station),
중고차매매단지 (used car sales complex), 지하상가 (underground shopping center), 휴게소 (rest area), 예식장 (wedding hall), 매표소 (ticket booth), 놀이시설 (amusement facility), 건물 (building), 모델하우스 (show house), 자동차매매단지 (car sales complex)

\paragraph{연구개발기관 (Research and Development Institutions):}
연구소 (research institute), 주행시험장 (driving test center)
\paragraph{산업\textbullet 물류 (Industrial Development and Logistic Complex):}
공단 (public corporation),  
물류단지 (logistics complex), 산업단지 (industrial complex)
\paragraph{복합단지 및 개발지구 (Industrial Development and Logistic Complex):} 
친환경복합단지 (eco-friendly complex)
\paragraph{금융관련공공기관 (financial regulators):} 
금융기관 (financial services agency) , 은행 (bank), 저축은행 (savings bank)

\subsubsection{사업체 (Corporate entities)}
\paragraph{외식업 (Eating and Drinking Places):}
가요주점 (karaoke pub), 고깃집 (Korean BBQ restaurant), 노래주점 (singing bar), 다방 (traditional Korean cafe), 동남아음식점 (Southeast Asian restaurant), 라이브카페 (live music cafe), 레스토랑 (restaurant), 바 (bar), 분식점 (snack bar), 뷔페 (buffet restaurant), 애견카페 (pet cafe), 일식당 (Japanese restaurant), 주점 (pub), 중식당 (Chinese restaurant), 치킨집 (fried chicken restaurant), 푸드트럭 (food truck), 한식당 (Korean restaurant), 해외식당 (international restaurant), 횟집 (sashimi restaurant), 키즈카페 (kids cafe), 카페 (cafe)

\paragraph{도\textbullet 소매 및 유통 (Wholesale and Retail Trade):}
가게 (shop), 가구매장 (furniture store), 가스업체 (gas supply company),
가전제품판매업 (home appliance store), 
고물업체 (scrap metal business), 
골프용품판매점 (golf equipment store), 과일가게 (fruit shop), 귀금속점 (jewelry store), 꽃가게 (flower shop), 농산물판매업 (agricultural product sales),  대리점 (distributor), 떡집 (rice cake shop), 마트 (grocery store), 문구점 (stationery store), 반찬가게 (side dish shop), 백화점 (department store), 빵집 (bakery), 상품권판매업체 (gift certificate vendor), 생활용품매장 (household goods store), 석유대체연료판매업체 (alternative fuel retailer), 슈퍼마켓 (supermarket), 스포츠용품점 (sporting goods store), 스포츠의류 (sportswear), 식품유통업 (food distribution business), 신발판매업체 (shoe store), 악기회사 (musical instrument company), 안경점 (optical shop), 반려동물분양업체 (pet shop), 약국 (pharmacy), 오디오샵 (audio equipment store), 옷가게 (clothing store), 원단공급업체 (fabric supplier), 유압벨브판매업체 (hydraulic valve vendor), 유통업 (distribution business), 의류매장 (apparel store), 자동차대리점 (car dealership), 자동차백화점 (auto megastore), 자동차판매점 (car sales shop), 전자제품매장 (electronics store), 정육점 (butcher shop), 제과점 (pastry shop), 주유소 (gas station), 
중고도서매매업체 (used book store), 
중고차매매업체 (used car dealership)
카드단말기판매업체 (credit card terminal distributor), 캠핑업체 (camping service provider), 컴퓨터판매업체 (computer retailer), 타이어판매업체 (tire shop),
페인트판매업 (paint supplier), 편의점 (convenience store), 화원 (flower shop), 화학약품판매업 (chemical supplier), 휴대전화판매업체 (mobile phone store), 휴대폰케이스매장 (mobile accessories shop), 보청기판매점 (hearing aid store), 자전거판매업 (bicycle shop), 기계도소매업 (machinery wholesale and retail business)
수산물유통업 (seafood distribution business), 매장 (store), 매점 (shop)

\paragraph{금융\textbullet세무 (Financial Institutions, Insurance and Other Financial Intermediaries):} 
, 
금융회사 (financial company), 대부업 (loan business), 보험사 (insuarance company), 신탁회사 (trust company), 전당포 (pawnshop), 증권사 (securities company), 카드회사 (credit card company), 투자회사 (investment frim), 해외은행 (foreign bank), 해외증권사 (foreign securities company), 세무법인 (tax corporation), 회계법인 (accounting corporation), 감정평가법인 (appraisal corporation), 집합투자기구 (collective investment scheme), 감정평가사사무소 (appraisal office), 세무사사무소 (tax accountant office)

\paragraph{법무 (Law Practice):} 
법률사무소 (law office), 법무법인 (law firm), 노무법인 (labor law firm), 법무사사무소 (judicial scrivener office)
\paragraph{부동산 중개 및 임대 매매 (Real Estate Business):}
공인중개사 (real estate agent), 부동산매매임대회사 (real estate sales and rental company), 부동산분양사무실 (real estate sales office), 분양대행사 (real estate marketing agency), 중개법인 (brokerage corporation)

\paragraph{정보통신업 (Information and Communications):}
방송국 (broadcasting station), 신문 (newspaper), 언론사 (media company), 출판사 (publishing company), 통신사 (telecommunications company), 전화국 (telephone office), 방송(broadcasting),

\paragraph{건설 (Construction):}
건설업체 (construction company), 부동산개발업 (real estate development business), 토공사 (civil engineering company), 토목업 (civil engineering business), 재개발업체 (redevelopment company), 조경업체 (landscaping company)

\paragraph{운수업 (Transportation):}
택배및운송회사 (transportation company), 택시회사 (taxi company), 여객운송회사 (passenger transport company), 이삿짐센터 (moving company)

\paragraph{물류 (Logistics and Distribution):}
물류센터 (logistics center), 물류창고 (logistics warehouse), 물류회사 (logistics company)

\paragraph{제조업 (Manufacturing):} 
가구공장 (furniture factory), 건설자재회사 (building materials company),
공장 (factory), 금속제조업 (metal manufacturing), 기계설비회사 (machinery and equipment company), 
목공소 (woodworking shop), 미용기기업체 (beauty equipment company), 보일러회사 (boiler manufacturer), 복합기업체 (multifunction printer manufacturer), 봉제업체 (sewing company), 비료회사 (fertilizer company), 석재가공업체 (stone processing company), 선박제조업 (shipbuilding company), 식품가공업 (food processing company), 식품업체(food company), 식품회사 (food company), 육류업체 (meat processing company), 
음료회사 (beverage company), 의료기기회사 (medical device company), 의류브랜드 (clothing brand), 이동식주택 (mobile home manufacturer), 
자동차부품생산업체 (auto parts manufacturer), 자동차회사 (automobile company), 전기배터리업체 (battery manufacturer), 전자전기제조업 (electronic component manufacturing), 
조선회사 (shipbuilding company), 주류회사 (alcoholic beverage company), 질소발생기제조업체 (nitrogen generator manufacturer), 철골제조업(steel frame manufacturing), 철판제조업 (steel plate manufacturing), 철판가공업 (sheet metal processing), 플라스틱가공업 (plastic processing company), 화장품회사 (cosmetics company), 중공업회사 (heavy industry company), 세라믹제조업 (ceramics manufacturing), 침장제조업 (bedding manufacturing company), 화학공업사 (chemical industry company), 정미소 (rice mill), 제약회사 (pharmaceutical company), 제철소 (steel mill)

\paragraph{농축물수산업 및 임업 (Agriculture, Fisheries, and Forestry):}
농장 (farm), 축산농장 (livestock farm), 어업회사 (fishery company), 과수원 (orchard)

\paragraph{광업 및 각종 자원 채굴\textbullet 채취 (Mining and Quarrying):}
금광채굴업 (gold mining), 광업소 (mining company)

\paragraph{숙박업 (Lodging and Accommodation):}
고시원 (gosiwon: a small single-room accommodation), 리조트 (resort), 모텔 (motel), 무인텔 (unmanned motel), 산장 (mountain lodge), 여관 (inn), 콘도 (condominium resort), 펜션 (pension), 호텔 (hotel), 게스트하우스 (guest house), 숙박시설 (lodging facility)

\paragraph{오락 및 스포츠 (Recreation, Leisure and Sports):}
극단 (theater troupe), 
PC방 (internet café), 게임장 (arcade), 골프연습장 (golf practice range), 골프장 (golf club), 낚시터 (fishing spot), 노래방 (karaoke room), 당구장 (billiard hall), 볼링장 (bowling alley), 수영장 (swimming pool), 승마장 (equestrian center), 실내낚시터 (indoor fishing cafe), 영화관 (movie theater), 오락실 (arcade), 온천 (hot spring), 워터파크 (water park), 캠핑장 (campground), 풀장 (pool), 헬스장 (fitness center), 기원 (baduk club), 당구장 (billiard hall), 수족관 (aquarium), 야구연습장 (batting cage), 스키장 (ski resort), 수상레저업(water leisure business)

\paragraph{미용\textbullet 욕탕\textbullet 신체관리 서비스 (Beauty and Body Care):}
마사지 (massage shop), 목욕탕 (public bathhouse), 미용실 (hair salon), 사우나 (sauna), 안마시술소 (massage parlor), 안마원 (therapeutic massage clinic), 왁싱샵 (waxing shop), 이발소 (barbershop), 찜질방 (Korean spa), 피어싱 (piercing studio), 반려동물미용샵 (pet grooming shop), 네일샵 (nail salon)

\paragraph{유흥업 (Adult entertainment):}
나이트클럽 (nightclub), 노래빠 (karaoke bar), 단란주점 (karaoke lounge with host services), 룸살롱 (high-end adult entertainment venue with private rooms), 업소 (adult entertainment venue), 카지노 (casino), 클럽 (club), 호스트바 (host bar), 무도장 (dance hall)

\paragraph{서비스 일반 (Other Service Sectors):}
건물임대관리회사 (building management company), 
광고회사 (advertising company),
대리운전회사 (designated driver service company), 동물병원 (veterinary clinic), 방역회사 (pest control company), 배달대행업체 (delivery agency), 상담소 (counseling center), 상조회사 (funeral service agency), 
선박임대판매업 (ship rental and sales company), 
세차장 (car wash), 세탁소 (laundry), 소개소 (labor dispatch agency), 스튜디오 (studio), 여행사 (travel agency), 오토바이수리점 (motorcycle repair shop), 요가학원 (yoga studio), 용역회사 (outsourcing service company), 운전면허학원 (driving school), 유학알선업체 (study abroad agency), 인터넷설치업체 (internet installation service), 인테리어 (interior design service), 
자동차임대업체 (car rental company), 재직정보제공업체 (employment verification company), 전자제품렌탈업 (electronics rental business), 정비공업사 (auto repair shop), 주차장 (parking lot), 주차장관리회사 (parking lot management company), 철거업체 (demolition company), 철학관 (fortune telling house), 청소대행업체 (cleaning service company), 컨설팅 (consulting firm), 태권도장 (Taekwondo gym), 택시면허매매중개업 (taxi license brokerage), 파티룸 (party room), 학원 (private academy), 기업행사대행업체 (event management company), 화실 (art studio), 점집 (fortune telling house), 운전면허시험장 (driver’s license test center), 보안경비회사 (security management company), 폐기물처리업체 (waste disposal company), 금속분석업 (metal analysis business), 산후조리원 (postnatal care center), 결혼준비대행업 (wedding planning agency), 건축사사무소 (architectural office), 사진관 (photo studio), 음원서비스 (music streaming service)

\paragraph{기업 일반 (Companies and Businesses in General):}
IT회사 (IT company), 불특정회사명 (unspecified company), 무역회사 (trading company), 유한공사 (limited company), 유한회사(limited liability company), 주식회사 (corporation), 지점 (branch office), 지주회사 (holding company), 국외기업 (foreign company), 상사회사 (trading company), 합자회사 (limited partnership company), 합명회사 (general partnership company), 농업회사법인 (agricultural corporation), 영농조합법인 (agricultural cooperative corporation)

\subsubsection{상품 일반 (Consumer Products)}
\paragraph{식\textbullet 의약품 (Foods and Medical Products):}
식품(food), 음료 (beverage), 의약품 (pharmaceutical product), 피자 (pizza)
\paragraph{공산품 (Industrial Products):}
가전제품 (home appliance), 공작기계 (machine tool), 마스크팩 (sheet mask), 발전기 (generator), 악기모델명 (instrument model name), 임플란트제품명 (dental implant product), 작업용차량 (work vehicle), 차량종류 (vehicle type), 철강제품 (steel product), 의료기기 (medical device), 교구 (teaching aid), 미용제품 (cosmetic product), 불특정제품명 (unspecified product name), 농약 (pesticide), 비료 (fertilizer), 항공기 (aircraft)

\paragraph{출판물 (Publications)}
서적 (books)

\paragraph{정보통신 상품 (Computer Equipment and Software):}
소프트웨어 (software)

\subsubsection{방송통신서비스 (Media and Telecommunications)}
\paragraph{온\textbullet 오프라인 방송 (Streaming and Broadcasting service):}
방송마일리지 (streaming donation points), 방송프로그램 (broadcasting program), 방송플랫폼 (streaming platform)
\paragraph{플랫폼 일반 (Online platforms in General):}
구인사이트 (job search site), 번역사이트 (translation site), 사이트 (uncategorized websites), 어플 (application), 포털 (portal site), 보이스피싱어플리케이션 (voice phising application)
\paragraph{전자상거래 (E-commerce):}
미술품경매사이트 (art auction site), 배달어플리케이션 (delivery application), 쇼핑몰 (online shopping mall), 중고거래사이트 (secondhand marketplace website)
\paragraph{소셜미디어 (Social Media):}
SNS (social networking service), 밴드 (group communication application), 소개팅어플리케이션 (dating application), 인터넷동성사이트 (LGBT dating app), 채팅어플리케이션 (chatting application), 커뮤니티사이트 (online community site), 국외메신저 (foreign messaging application), 온라인게시판명 (name of online bulletin board), 온라인게시글명 (title of online post), 온라인대화방명 (name of online chat room)

\paragraph{게임 (Online Games):}
게임마일리지 (game mileage), 게임서버 (game server), 게임아이템 (in-game item), 게임아이템거래카페 (item trading forum), 모바일게임 (mobile game), 온라인게임 (online game), 인터넷도박 (online gambling)

\subsubsection{금융서비스 (Financial Products and Services)}
\paragraph{투자\textbullet 보험\textbullet 대출 서비스 (Investment, Insurance and Personal Loan Services):}
골프보험 (golf insurance), 금융투자상품 (financial investment product), 대출상품 (loan product), 보험상품 (insurance plan)
\paragraph{가상자산 (Virtual Assets):}
가상화폐 (cryptocurrency), 가상화폐거래프로그램 (crypto trading platform), 가상화폐거래소 (cryptocurrency exchange)

\subsubsection{사회\textbullet 문화 (Culture and Society)}
\paragraph{국가유산 (National Heritage and Other Cultural Features):} 
중요무형문화재 (intangible cultural heritage) 
\paragraph{예술 (Fine Arts, Visual Arts, Performing Arts):}
공연 (performance), 영화 (film)
\paragraph{교육 및 학술 (Education Programs and Academic Curriculum):}
교과목 (curriculum)
\paragraph{각종 행사 (Socio-cultural Events):}
공청회 (public hearing), 낚시대회 (fishing competition), 등산행사 (hiking event), 임플란트세미나 (implant seminar), 사회공헌\textbullet자선행사 (charity event), 축제 (festival), 행사 (uncategorized events)
\paragraph{스포츠 (Sports)}
운동종목 (sports category)
\paragraph{각종 과업 (Various Projects)}
공사 (construction work), 사업 (project), 용역 (service contract)
\subsubsection{URL}
\paragraph{URL:} URL

\section{Model and Training}
\label{app:training}

This section provides additional details on our model architecture, and training procedures introduced in the main paper (Section~\ref{sec:experiments}). We first describe \sys model family developed for Korean court judgment de-identification. We then outline the pre-training and fine-tuning strategies applied to both our models and the baselines. 

\begin{table*}[b]
\begin{adjustbox}{max width=\textwidth}
    \centering
    \small
    \begin{tabular}{lccccc}
    \toprule
    \textbf{Aspect} & \textbf{\sys-370M} & \textbf{\sys-800M} & \textbf{\sys-1.5B} & \textbf{Polyglot-Ko} & \textbf{EXAONE-3.5} \\
    \midrule
    Parameters & 370M & 800M & 1.5B & 1.3B & 2.4B \\
    Hidden Dimension & 1024 & 1280 & 2048 & 2048 & 2560 \\
    Transformer Layers & 24 & 36 & 24 & 24 & 30 \\
    Attention Heads & 16 & 20 & 32 & 16 & 32 \\
    Vocabulary Size & 32,000 & 32,000 & 128,000 & 30,080 & 102,400 \\
    \midrule
    Pre-train Corpus & 14B (7B Ko / 7B En) & 30B (15B Ko / 15B En) & 60B (22B Ko / 38B En) & - & - \\
    Pre-train Hardware & 32× NVIDIA H100 80GB & 32× NVIDIA H100 80GB & 32× NVIDIA H100 80GB & - & - \\
    Pre-train Duration & 2 hours & 9 hours & 19 hours & - & - \\
    Pre-train Learning Rate & 7.5e-5 & 7.5e-5 & 7.5e-5 & - & - \\
    Pre-train Batch Size & 2048 & 2048 & 2048 & - & - \\
    Pre-train Seq Length & 512 $\rightarrow$ 2048 & 512 $\rightarrow$ 2048 & 512 $\rightarrow$ 2048 & - & - \\
    Pre-train AdamW Betas & $\beta = (0.9, 0.98)$ & $\beta = (0.9, 0.98)$ & $\beta = (0.9, 0.98)$ & - & - \\
    Pre-train AdamW Weight Decay & 0.01 & 0.01 & 0.01 & - & - \\
    \midrule
    Fine-tuning Hardware & 8× NVIDIA H100 80GB & 8× NVIDIA H100 80GB & 8× NVIDIA H100 80GB & 8× NVIDIA H100 80GB & 8× NVIDIA H100 80GB \\
    Fine-tuning Learning Rate & 5e-5 & 2e-5 & 2e-5 & 2e-5 & 2e-5 \\
    Fine-tuning Batch Size & 32 & 32 & 32 & 32 & 32 \\
    Fine-tuning Seq Length & 2048 & 2048 & 2048 & 2048 & 2048 \\
    Fine-tuning AdamW Betas & $\beta = (0.9, 0.98)$ & $\beta = (0.9, 0.98)$ & $\beta = (0.9, 0.98)$ & $\beta = (0.9, 0.98)$ & $\beta = (0.9, 0.98)$ \\
    Fine-tuning AdamW Weight Decay & 0.01 & 0.01 & 0.01 & 0.01 & 0.01 \\
    \bottomrule
    \end{tabular}
    \end{adjustbox}
    \caption{Comparison of \sys models and baseline Korean models, Exaone and Polyglot-ko.}
    \label{tab:model_config}
\end{table*}

\subsection{Model configuration}
\paragraph{Model.} We introduce \sys, a family of models based on the DeBERTa-v3 architecture, designed for de-identification through token classification. \sys family includes three models: 370M, 800M and 1.5B models. The 370M model has 370 million parameters, a hidden dimension of 1024, 24 transformer layers, 16 attention heads, and a vocabulary size of 32,000. The 800M model has 800 million parameters, a hidden dimension of 1280, 36 transformer layers, 20 attention heads, and a vocabulary size of 32,000. The 1.5B model has 1.5 billion parameters, a hidden dimension of 2048, 24 transformer layers, 32 attention heads, and a vocabulary size of 128,000. The smaller vocabulary size for the 370M and 800M models prevents the embedding matrix from becoming disproportionately large relative to the transformer layers to ensure balanced model architecture. 


\subsection{Training}
\paragraph{Pre-training.} \sys models are pre-trained from scratch on the bilingual corpus from Section 4.1, which yields 60 billion tokens (22 billion Korean, 38 billion English) when tokenized with our custom tokenizer. The 370M model is pre-trained on a 14 billion token subset (7 billion Korean, 7 billion English) sampled from the corpus, conducted over 2 hours using 32 NVIDIA H100 80GB GPUs. The 800M model is pre-trained on a 30 billion token subset (15 billion Korean, 15 billion English) sampled from the corpus, conducted over 9 hours using 32 NVIDIA H100 80GB GPUs. The 1.5B model is pre-trained on the full 60 billion tokens (22 billion Korean, 38 billion English), conducted over 19 hours using 32 NVIDIA H100 80GB GPUs. For the 370M model, initial pre-training uses a global batch size of 2048, a peak learning rate of 7.5e-5, a masked language modeling (MLM) probability of 0.15, and a maximum sequence length of 512, with the DeepSpeed framework under ZeRO Stage 0 (DDP). For the 800M and 1.5B models, the same configuration is used but with a peak learning rate of 5e-5. All models are optimized using AdamW~\cite{adamw} optimizer with $\beta = (0.9, 0.999)$. A learning rate schedule with a warm-up phase for the first 10\% of training steps and cosine decay for the remainder is applied across all models. To handle longer inputs, each model undergoes additional training on 2 million tokens with a maximum sequence length of 2048, using the same learning rate schedule. All models use FP16 mixed precision~\cite{micikevicius2017mixed} training.

\paragraph{Fine-Tuning.} We fine-tune \sys models and the baseline models on the token classification task using the dataset (Section 4.1). We employ two data augmentation settings: Per-Epoch Entity Replacement, where entity mentions in each document are replaced with new samples from a predefined list at every epoch to increase data diversity, and Single Replacement, where entity mentions are replaced once and remain fixed throughout training. At each epoch under Per-Epoch Entity Replacement, the model sees a different variant of every document, and the full training completes over 30 epochs to cover the entire augmented set. The validation set remains unchanged to ensure consistent evaluation. For the 370M model, we set the global batch size to 32, the peak learning rate to 5e-5. For the 800M model, 1.5B model, Polyglot-Ko and Exaone-3.5, the same configuration is used but with a peak learning rate of 2e-5. All models are trained with a maximum input length capped at 2048 tokens (the model limit). Inputs longer than this limit are truncated from the end (head-only, tail truncation), so very long court rulings—especially civil and administrative cases—may not be fully covered by the model input. We apply FP16 mixed precision across all models and optimize these models using AdamW optimizer with $\beta = (0.9, 0.999)$.
\section{Evaluation Metrics}
\label{app:metrics}
This section details the evaluation metrics used to assess model performance in the de-identification of Korean court judgments, as discussed in the main paper (Section \ref{sec:experiments}). We describe Binary Token-Level F1 and Token-Level Micro F1, including their mathematical definitions and significance for result analysis.

\paragraph{Backgrounds.}
In token classification for de-identification, model performance is measured with true positives (TP), false positives (FP), and false negatives (FN). TP is the number of tokens correctly predicted as a target label. FP is the number of tokens incorrectly predicted as a target label when they belong to another label. FN is the number of tokens belonging to a target label but incorrectly predicted as another label. Precision is the proportion of correctly predicted tokens among all tokens predicted as the target label, and recall is the proportion of correctly predicted tokens among all tokens truly belonging to the target label. These are defined as:\vspace{1mm} 
\begin{align*}
\small
\text{Precision} &= 
\small\frac{\text{TP}}{\text{TP} + \text{FP}} \\
\small\text{Recall} &= 
\small\frac{\text{TP}}{\text{TP} + \text{FN}}
\end{align*}

\paragraph{Binary Token-Level F1.} Binary Token-Level F1 evaluates the model’s ability to classify tokens requiring de-identification from those that do not regardless of entity type. High scores ensure accurate detection of all tokens requiring de-identification like “\scalebox{0.9}{홍길동}” (Hong Gildong) while excluding others like “\scalebox{0.9}{이}” (i). This metric is critical because missing even one token that requires de-identification can immediately lead to increase identifiability of the person and thus compromise privacy. By treating all entity types as a single class, it provides a simple yet robust baseline widely adopted in de-identification research~\cite{dernoncourt2016deid, yuezhou2020phicon, salierno2024giusberto, kim2024generalizing}. 
The binary token-level F1 score in our experiment is calculated as follows:\vspace{1mm} 
\begin{align*}
\small
\text{Binary Precision} &= \small\frac{\text{TP}_{\text{bin}}}{\text{TP}_{\text{bin}} + \text{FP}_{\text{bin}}} \\
\small
\text{Binary Recall} &= \small\frac{\text{TP}_{\text{bin}}}{\text{TP}_{\text{bin}} + \text{FN}_{\text{bin}}}
\end{align*}
\begin{equation*}
\text{\(
\substack{\text{Binary Token-Level} \\ \text{F1}}
\)} 
\small= 2 \cdot \frac{\text{Binary Precision} \cdot \text{Binary Recall}}{\text{Binary Precision} + \text{Binary Recall}}
\end{equation*}\vspace{5mm}

where the positive class is any non-“Outside” label (e.g., name, phone number). Here, $\text{TP}_{\text{bin}}$ is the number of tokens truly non-“Outside” and correctly predicted as non-“Outside”, $\text{FP}_{\text{bin}}$ is the number of tokens actually “Outside” but incorrectly predicted as non-“Outside”, and $\text{FN}_{\text{bin}}$ is the number of tokens truly non-“Outside” but incorrectly predicted as “Outside”.

\paragraph{Token-Level Micro F1.} Token-Level Micro F1 measures how well the model classifies tokens into specific entity types such as name of the person and phone numbers. It excludes the “Outside” label and calculates performance using aggregated precision and recall for each entity type. High scores indicate correct identification and labeling of tokens requiring de-identification, such as classifying “\scalebox{0.9}{홍길동}” (Hong Gildong) as a name of the person rather than a corporate entity. 

Accurate classification of entity types is essential for proper de-identification of court judgments. This gets importance in the post-processing stage because without precise entity type prediction, the identified parts containing sensitive information cannot be properly replaced with contextually congruent phrases. Inaccurate classification can result in awkward or incorrect replacements in post-processing and ultimately lead to undermine the readability of the anonymized text. 

For example, account numbers must be accurately identified and replaced with phrases like “\scalebox{0.9}{계좌번호} 1 \scalebox{0.9}{생략}” (Account number 1 omitted) during post-processing. If classified as different entity type such as a phone number, the misclassified account number might be incorrectly replaced with a phrase like “\scalebox{0.9}{전화번호} 1 \scalebox{0.9}{생략}” (Phone number 1 omitted). If the same case is classified as a business entity, it might be replaced with “A”, and the post-processing result is not compatible with the current law and practice concerning the methods of anonymization (Judicial Rule No. 1778).

The token-level F1 score in our experiment is calculated as follows:
\begin{align*}
\small\text{Micro Precision} &= \frac{\sum_{c \in C} \text{TP}_c}{\sum_{c \in C} (\text{TP}_c + \text{FP}_c)} \\
\small\text{Micro Recall} &= \frac{\sum_{c \in C} \text{TP}_c}{\sum_{c \in C} (\text{TP}_c + \text{FN}_c)}
\end{align*}
\begin{equation*}
\text{\(
\substack{\text{Token-Level} \\ \text{Micro F1}}
\)} 
\small= 2 \cdot \frac{\text{Micro Precision} \cdot \text{Micro Recall}}{\text{Micro Precision} + \text{Micro Recall}}
\end{equation*}\vspace{5mm}

where $C$ is the set of entity types (labels excluding the “Outside”), and for each entity type $c \in C$, $\text{TP}_c$, $\text{FP}_c$, and $\text{FN}_c$ are the true positives, false positives, and false negatives respectively.

\section{Annotators}
The authors participated in the annotation process for 20 hours per week over a period of 4 weeks. Seventeen external annotators contributed to the task for 12 hours per week over 4 weeks. These annotators were compensated at a rate of 10,000 KRW per hour, amounting to a total payment of 480,000 KRW per person. We consider this compensation appropriate given the local standards of living and the scope of the work.

\section{Performance by Case Type}
\label{app:case-results}
We report case-type precision, recall, binary token-level F1, and token-level micro F1 under two data regimes: Single Replacement and Per-Epoch Entity Replacement discussed in ~\ref{sec:exp-results}.

See the tables below for detailed results: binary token-level in \autoref{tab:case-bin-single} (Single) and \autoref{tab:case-bin-per-epoch} (Per-Epoch), and token-level in \autoref{tab:case-token-single} (Single) and \autoref{tab:case-token-per-epoch} (Per-Epoch). All tables report Precision and Recall; F1 is binary for binary token-level and micro-averaged for token-level. All values are averaged over three runs (seeds 1200, 1203, 1205) for each case type.


\begin{table*}[!t]
\small
\centering
\resizebox{0.9\linewidth}{!}{%

\begin{tabular}{l l l *{3}{>{\centering\arraybackslash}m{1.6cm}}}
\toprule
\multirow{2}{*}{\textbf{Domain}} & \multirow{2}{*}{\textbf{Case type}} &
\multirow{2}{*}{\textbf{Model}} &
\multicolumn{3}{c}{\makecell[c]{\textbf{Single Replacement}\\\textbf{(Binary Token-Level)}}} \\
\cmidrule(lr){4-6}
& & & \textbf{P} & \textbf{R} & \textbf{F1} \\
\midrule
\multirow{20}{*}{Civil} & \multirow{5}{*}{\makecell[l]{Compensation\\for damage}} 
  & Polyglot-ko (1.3B) & 0.9843 & 0.9589 & 0.9714 \\
& & Exaone (2.4B)      & 0.9818 & 0.9462 & 0.9637 \\
& & \sys-360M    & 0.9718 & 0.9303 & 0.9506 \\
& & \sys-800M    & 0.9870 & 0.9774 & 0.9822 \\
& & \sys-1.5B    & 0.9954 & 0.9663 & 0.9806 \\
\cmidrule(lr){2-6}
& \multirow{5}{*}{Eviction} 
  & Polyglot-ko (1.3B) & 0.9700 & 0.9336 & 0.9514 \\
& & Exaone (2.4B)     & 0.9681 & 0.9529 & 0.9604  \\
& & \sys-360M    & 0.9672 & 0.9070 & 0.9361 \\
& & \sys-800M    & 0.9763 & 0.9506 & 0.9632 \\
& & \sys-1.5B    & 0.9709 & 0.9632 & 0.9671 \\
\cmidrule(lr){2-6}
& \multirow{5}{*}{\makecell[l]{Purchase-price\\ of a sale}} 
  & Polyglot-ko (1.3B) & 0.9727 & 0.9520 & 0.9623 \\
& & Exaone (2.4B)     & 0.9812 & 0.9594 & 0.9701 \\
& & \sys-360M    &  0.9713 & 0.9148 & 0.9421 \\
& & \sys-800M    & 0.9851 & 0.9628 & 0.9738\\
& & \sys-1.5B    & 0.9854 & 0.9600 & 0.9725 \\
\cmidrule(lr){2-6}
& \multirow{5}{*}{\makecell[l]{Security deposit\\disputes}} 
  & Polyglot-ko (1.3B) & 0.9865 & 0.9726 & 0.9795 \\
& & Exaone (2.4B)     & 0.9826 & 0.9614 & 0.9719 \\
& & \sys-360M    & 0.9816 & 0.9317 & 0.9559\\
& & \sys-800M    & 0.9865 & 0.9684 & 0.9773 \\
& & \sys-1.5B    & 0.9881 & 0.9695 & 0.9787 \\
\midrule
\multirow{25}{*}{Criminal} & \multirow{5}{*}{Bodily injury} 
  & Polyglot-ko (1.3B) & 0.9826 & 0.9588 & 0.9706 \\
& & Exaone (2.4B)      & 0.9724 & 0.9718 & 0.9720\\
& & \sys-360M    &   0.9886 & 0.9623 & 0.9752  \\
& & \sys-800M    & 0.9905 & 0.9800 & 0.9852  \\
& & \sys-1.5B    & 0.9884 & 0.9806 & 0.9845 \\
\cmidrule(lr){2-6}
& \multirow{5}{*}{Drunk driving} 
  & Polyglot-ko (1.3B) & 0.9728 & 0.9508 & 0.9616 \\
& & Exaone (2.4B)       & 0.9831 & 0.9473 & 0.9649 \\
& & \sys-360M    & 0.9714 & 0.9164 & 0.9430 \\
& & \sys-800M    & 0.9733 & 0.9488 & 0.9608 \\
& & \sys-1.5B    & 0.9817 & 0.9508 & 0.9660 \\
\cmidrule(lr){2-6}
& \multirow{5}{*}{\makecell{Property theft\\and deception}} 
  & Polyglot-ko (1.3B) & 0.9775 & 0.9659 & 0.9717 \\
& & Exaone (2.4B)      & 0.9707 & 0.9601 & 0.9654\\
& & \sys-360M    & 0.9845 & 0.9418 & 0.9627  \\
& & \sys-800M    & 0.9718 & 0.9806 & 0.9762 \\
& & \sys-1.5B    & 0.9911 & 0.9766 & 0.9838 \\
\cmidrule(lr){2-6}
& \multirow{5}{*}{Sexual misconduct} 
  & Polyglot-ko (1.3B) & 0.9837 & 0.9690 & 0.9763 \\
& & Exaone (2.4B)      & 0.9837 & 0.9561 & 0.9697\\
& & \sys-360M    & 0.9803 & 0.9260 & 0.9524\\
& & \sys-800M    & 0.9872 & 0.9705 & 0.9788 \\
& & \sys-1.5B   & 0.9881 & 0.9650 & 0.9764 \\
\cmidrule(lr){2-6}
& \multirow{5}{*}{Violence} 
  & Polyglot-ko (1.3B) & 0.9758 & 0.9644 & 0.9701 \\
& & Exaone (2.4B)      & 0.9702 & 0.9701 & 0.9701\\
& & \sys-360M    & 0.9664 & 0.9316 & 0.9486 \\
& & \sys-800M    & 0.9749 & 0.9778 & 0.9763 \\
& & \sys-1.5B    & 0.9740 & 0.9809 & 0.9774 \\
\midrule
\multirow{5}{*}{Administrative} & \multirow{5}{*}{\makecell[l]{Administrative\\litigation}} 
  & Polyglot-ko (1.3B) & 0.9641 & 0.9321 & 0.9478 \\
& & Exaone (2.4B)      & 0.9743 & 0.9383 & 0.9559 \\
& & \sys-360M    &0.9814 & 0.9254 & 0.9526 \\
& & \sys-800M    & 0.9877 & 0.9555 & 0.9713 \\
& & \sys-1.5B    & 0.9842 & 0.9811 & 0.9827 \\
\bottomrule
\end{tabular}}

\caption{Binary token-level metrics (Precision, Recall, and F1) for the \textbf{Single Replacement} setting, reported by case type and model (parameters shown in parentheses).}

\label{tab:case-bin-single}
\end{table*}


\newcolumntype{C}{>{\centering\arraybackslash}p{1.5cm}}

\begin{table*}[!t]
\small
\centering
\resizebox{0.9\linewidth}{!}{%

\begin{tabular}{l l l *{3}{>{\centering\arraybackslash}m{1.6cm}}}
\toprule
\multirow{2}{*}{\textbf{Domain}} & \multirow{2}{*}{\textbf{Case type}} &
\multirow{2}{*}{\textbf{Model}} &
\multicolumn{3}{c}{\makecell[c]{\textbf{Per-Epoch Replacement}\\\textbf{(Binary Token-Level)}}} \\
\cmidrule(lr){4-6}
& & & \textbf{P} & \textbf{R} & \textbf{F1} \\
\midrule
\multirow{20}{*}{Civil} & \multirow{5}{*}{\makecell[l]{Compensation\\for damage}} 
  & Polyglot-ko (1.3B) & 0.9779 & 0.9687 & 0.9732           \\
& & Exaone (2.4B)      & 0.9770 & 0.9591 & 0.9679 \\
& & \sys-360M    & 0.9611 & 0.9763 & 0.9686   \\
& & \sys-800M    & 0.9796 & 0.9889 & 0.9842          \\
& & \sys-1.5B    & 0.9796 & 0.9891 & 0.9843         \\
\cmidrule(lr){2-6}
& \multirow{5}{*}{Eviction} 
  & Polyglot-ko (1.3B) & 0.9644 & 0.9639 & 0.9641          \\
& & Exaone (2.4B)      &  0.9635 & 0.9597 & 0.9616 \\
& & \sys-360M    & 0.9482 & 0.9566 & 0.9524  \\
& & \sys-800M    & 0.9615 & 0.9711 & 0.9663           \\
& & \sys-1.5B    & 0.9569 & 0.9803 & 0.9685           \\
\cmidrule(lr){2-6}
& \multirow{5}{*}{\makecell[l]{Payment of\\purchase price}} 
  & Polyglot-ko (1.3B) & 0.9630 & 0.9650 & 0.9640          \\
& & Exaone (2.4B)      & 0.9679 & 0.9714 & 0.9696\\
& & \sys-360M    & 0.9463 & 0.9667 & 0.9564   \\
& & \sys-800M    & 0.9712 & 0.9822 & 0.9766          \\
& & \sys-1.5B    & 0.9748 & 0.9851 & 0.9799           \\
\cmidrule(lr){2-6}
& \multirow{5}{*}{\makecell[l]{Security deposit\\disputes}} 
  & Polyglot-ko (1.3B)& 0.9770 & 0.9732 & 0.9751           \\
& & Exaone (2.4B)      & 0.9732 & 0.9736 & 0.9734 \\
& & \sys-360M    & 0.9714 & 0.9661 & 0.9687   \\
& & \sys-800M    & 0.9795 & 0.9864 & 0.9829           \\
& & \sys-1.5B   & 0.9807 & 0.9878 & 0.9842          \\
\midrule
\multirow{25}{*}{Criminal} & \multirow{5}{*}{Bodily injury} 
  & Polyglot-ko (1.3B) & 0.9777 & 0.9746 & 0.9761           \\
& & Exaone (2.4B)      &  0.9697 & 0.9803 & 0.9749 \\
& & \sys-360M    & 0.9811 & 0.9820 & 0.9815 \\
& & \sys-800M    & 0.9875 & 0.9870 & 0.9872          \\
& & \sys-1.5B    & 0.9868 & 0.9898 & 0.9883           \\
\cmidrule(lr){2-6}
& \multirow{5}{*}{Drunk driving} 
  & Polyglot-ko (1.3B) & 0.9667 & 0.9645 & 0.9656 \\
& & Exaone (2.4B)      & 0.9726 & 0.9685 & 0.9705\\
& & \sys-360M    & 0.9612 & 0.9572 & 0.9592           \\
& & \sys-800M    & 0.9592 & 0.9795 & 0.9692          \\
& & \sys-1.5B    & 0.9660 & 0.9739 & 0.9699           \\
\cmidrule(lr){2-6}
& \multirow{5}{*}{\makecell{Property theft\\and deception}} 
  & Polyglot-ko (1.3B) & 0.9754 & 0.9776 & 0.9765           \\
& & Exaone (2.4B)      & 0.9651 & 0.9702 & 0.9676\\
& & \sys-360M    & 0.9739 & 0.9767 & 0.9753           \\
& & \sys-800M    & 0.9843 & 0.9840 & 0.9841           \\
& & \sys-1.5B   & 0.9850 & 0.9895 & 0.9873           \\
\cmidrule(lr){2-6}
& \multirow{5}{*}{Sexual misconduct} 
  & Polyglot-ko (1.3B) & 0.9788 & 0.9744 & 0.9766           \\
& & Exaone (2.4B)     & 0.9770 & 0.9638 & 0.9705\\
& & \sys-360M    & 0.9667 & 0.9698 & 0.9682           \\
& & \sys-800M    & 0.9814 & 0.9840 & 0.9827           \\
& & \sys-1.5B    & 0.9786 & 0.9851 & 0.9818           \\
\cmidrule(lr){2-6}
& \multirow{5}{*}{Violence} 
  & Polyglot-ko (1.3B) & 0.9679 & 0.9736 & 0.9707           \\
& & Exaone (2.4B)      & 0.9610 & 0.9754 & 0.9681 \\
& & \sys-360M    & 0.9599 & 0.9745 & 0.9672           \\
& & \sys-800M    & 0.9706 & 0.9874 & 0.9789           \\
& & \sys-1.5B   & 0.9724 & 0.9942 & 0.9831           \\
\midrule
\multirow{5}{*}{Administrative} & \multirow{5}{*}{\makecell[l]{Administrative\\litigation}}
  & Polyglot-ko (1.3B) & 0.9603 & 0.9564 & 0.9583  \\
& & Exaone (2.4B)      & 0.9589 & 0.9605 & 0.9597\\
& & \sys-360M    & 0.9666 & 0.9623 & 0.9644          \\
& & \sys-800M    & 0.9802 & 0.9814 & 0.9808           \\
& & \sys-1.5B    & 0.9739 & 0.9898 & 0.9818           \\
\bottomrule
\end{tabular}}

\caption{Binary token-level metrics (Precision, Recall, and F1) for the \textbf{Per-Epoch Replacement} setting, reported by case type and model (parameters shown in parentheses).}

\label{tab:case-bin-per-epoch}
\end{table*}


\begin{table*}[!t]
\small
\centering
\resizebox{0.9\linewidth}{!}{%

\begin{tabular}{l l l *{3}{>{\centering\arraybackslash}m{1.6cm}}}
\toprule
\multirow{2}{*}{\textbf{Domain}} & \multirow{2}{*}{\textbf{Case type}} &
\multirow{2}{*}{\textbf{Model}} &
\multicolumn{3}{c}{\makecell[c]{\textbf{Single Replacement}\\\textbf{(Token-Level)}}} \\
\cmidrule(lr){4-6}
& & & \textbf{P} & \textbf{R} & \textbf{Micro F1} \\
\midrule
\multirow{20}{*}{Civil} & \multirow{5}{*}{\makecell[l]{Compensation\\for damage}} 
  & Polyglot-ko (1.3B) & 0.8793 & 0.8566 & 0.8677 \\
& & Exaone (2.4B)      & 0.8285 & 0.7988 & 0.8134 \\
& & \sys-360M    & 0.7518 & 0.7195 & 0.7352 \\
& & \sys-800M    & 0.7949 & 0.7872 & 0.7910 \\
& & \sys-1.5B    & 0.8280 & 0.8037 & 0.8156 \\
\cmidrule(lr){2-6}
& \multirow{5}{*}{Eviction} 
  & Polyglot-ko (1.3B) & 0.8936 & 0.8602 & 0.8766 \\
& & Exaone (2.4B)      & 0.9108 & 0.8965 & 0.9036 \\
& & \sys-360M    & 0.8963 & 0.8405 & 0.8675 \\
& & \sys-800M    & 0.9234 & 0.8989 & 0.9109 \\
& & \sys-1.5B    & 0.8985 & 0.8913 & 0.8949 \\
\cmidrule(lr){2-6}
& \multirow{5}{*}{\makecell[l]{Payment of\\purchase price}} 
  & Polyglot-ko (1.3B) & 0.8386 & 0.8207 & 0.8296 \\
& & Exaone (2.4B)     & 0.8189 & 0.8000 & 0.8092 \\
& & \sys-360M    & 0.8619 & 0.8118 & 0.8361 \\
& & \sys-800M    & 0.9057 & 0.8854 & 0.8954 \\
& & \sys-1.5B    & 0.9094 & 0.8859 & 0.8975 \\
\cmidrule(lr){2-6}
& \multirow{5}{*}{\makecell[l]{Security deposit\\disputes}} 
  & Polyglot-ko (1.3B) & 0.8991 & 0.8864 & 0.8927 \\
& & Exaone (2.4B)     & 0.8959 & 0.8766 & 0.8861 \\
& & \sys-360M    & 0.9312 & 0.8839 & 0.9069 \\
& & \sys-800M    & 0.9411 & 0.9239 & 0.9324 \\
& & \sys-1.5B    & 0.9440 & 0.9261 & 0.9349 \\
\midrule
\multirow{25}{*}{Criminal} & \multirow{5}{*}{Bodily injury} 
  & Polyglot-ko (1.3B) & 0.8852 & 0.8639 & 0.8744 \\
& & Exaone (2.4B)      & 0.8962 & 0.8956 & 0.8958 \\
& & \sys-360M    & 0.9344 & 0.9096 & 0.9218 \\
& & \sys-800M    & 0.9479 & 0.9378 & 0.9428 \\
& & \sys-1.5B    & 0.9433 & 0.9360 & 0.9396 \\
\cmidrule(lr){2-6}
& \multirow{5}{*}{Drunk driving} 
  & Polyglot-ko (1.3B) & 0.8644 & 0.8448 & 0.8545 \\
& & Exaone (2.4B)       & 0.9047 & 0.8718 & 0.8879 \\
& & \sys-360M    & 0.8784 & 0.8286 & 0.8527 \\
& & \sys-800M    & 0.9097 & 0.8867 & 0.8980 \\
& & \sys-1.5B    & 0.9078 & 0.8790 & 0.8931 \\
\cmidrule(lr){2-6}
& \multirow{5}{*}{\makecell{Property theft\\and deception}} 
  & Polyglot-ko (1.3B) & 0.9040 & 0.8933 & 0.8987  \\
& & Exaone (2.4B)      & 0.9039 & 0.8940 & 0.8989 \\
& & \sys-360M    & 0.9385 & 0.8978 & 0.9177 \\
& & \sys-800M    & 0.9456 & 0.9286 & 0.9370 \\
& & \sys-1.5B    & 0.9322 & 0.9186 & 0.9253  \\
\cmidrule(lr){2-6}
& \multirow{5}{*}{Sexual misconduct} 
  & Polyglot-ko (1.3B) & 0.8900 & 0.8767 & 0.8833 \\
& & Exaone (2.4B)      & 0.8505 & 0.8265 & 0.8383 \\
& & \sys-360M    & 0.8879 & 0.8387 & 0.8626 \\
& & \sys-800M    & 0.8958 & 0.8807 & 0.8882 \\
& & \sys-1.5B    & 0.8941 & 0.8731 & 0.8834 \\
\cmidrule(lr){2-6}
& \multirow{5}{*}{Violence} 
  & Polyglot-ko (1.3B) & 0.8704 & 0.8602 & 0.8653  \\
& & Exaone (2.4B)      & 0.8828 & 0.8826 & 0.8827 \\
& & \sys-360M    & 0.9036 & 0.8711 & 0.8871 \\
& & \sys-800M    & 0.9203 & 0.9231 & 0.9217 \\
& & \sys-1.5B    & 0.9138 & 0.9205 & 0.9171 \\
\midrule
\multirow{5}{*}{Administrative} & \multirow{5}{*}{\makecell[l]{Administrative\\litigation}} 
  & Polyglot-ko (1.3B) & 0.8661 & 0.8373 & 0.8515 \\
& & Exaone (2.4B)      & 0.8999 & 0.8666 & 0.8829 \\
& & \sys-360M    & 0.9246 & 0.8718 & 0.8974 \\
& & \sys-800M    & 0.9481 & 0.9172 & 0.9324 \\
& & \sys-1.5B    & 0.9363 & 0.9334 & 0.9349 \\
\bottomrule
\end{tabular}}

\caption{Token-level metrics (Precision, Recall, and Micro F1) for the \textbf{Single Replacement} setting, reported by case type and model (parameters shown in parentheses).}

\label{tab:case-token-single}
\end{table*}

\begin{table*}[!t]
\small
\centering
\resizebox{0.9\linewidth}{!}{%

\begin{tabular}{l l l *{3}{>{\centering\arraybackslash}m{1.6cm}}}
\toprule
\multirow{2}{*}{\textbf{Domain}} & \multirow{2}{*}{\textbf{Case type}} &
\multirow{2}{*}{\textbf{Model}} &
\multicolumn{3}{c}{\makecell[c]{\textbf{Per-Epoch Replacement}\\\textbf{(Token-Level)}}} \\
\cmidrule(lr){4-6}
& & & \textbf{P} & \textbf{R} & \textbf{Micro F1} \\
\midrule
\multirow{20}{*}{Civil} & \multirow{5}{*}{\makecell[l]{Compensation\\for damage}} 
  & Polyglot-ko (1.3B) & 0.8774 & 0.8688 & 0.8730 \\
& & Exaone (2.4B)      & 0.8525 & 0.8372 & 0.8448 \\
& & \sys-360M    & 0.7435 & 0.7553 & 0.7493 \\
& & \sys-800M    & 0.8121 & 0.8197 & 0.8159 \\
& & \sys-1.5B    & 0.8141 & 0.8220 & 0.8179  \\
\cmidrule(lr){2-6}
& \multirow{5}{*}{Eviction} 
  & Polyglot-ko (1.3B) & 0.8878 & 0.8874 & 0.8875 \\
& & Exaone (2.4B)     & 0.9007 & 0.8971 & 0.8989  \\
& & \sys-360M    & 0.8939 & 0.9019 & 0.8979 \\
& & \sys-800M    & 0.9035 & 0.9125 & 0.9080 \\
& & \sys-1.5B     & 0.8967 & 0.9186 & 0.9075 \\
\cmidrule(lr){2-6}
& \multirow{5}{*}{\makecell[l]{Payment of\\purchase price}} 
  & Polyglot-ko (1.3B) & 0.8322 & 0.8339 & 0.8330 \\
& & Exaone (2.4B)     & 0.8460 & 0.8490 & 0.8474 \\
& & \sys-360M    & 0.8646 & 0.8833 & 0.8738 \\
& & \sys-800M    & 0.8902 & 0.9002 & 0.8952 \\
& & \sys-1.5B    & 0.8933 & 0.9029 & 0.8981 \\
\cmidrule(lr){2-6}
& \multirow{5}{*}{\makecell[l]{Security deposit\\disputes}} 
  & Polyglot-ko (1.3B) & 0.8958 & 0.8923 & 0.8940 \\
& & Exaone (2.4B)    & 0.8833 & 0.8838 & 0.8835 \\
& & \sys-360M   & 0.9268 & 0.9218 & 0.9243\\
& & \sys-800M    & 0.9430 & 0.9497 & 0.9463 \\
& & \sys-1.5B   & 0.9469 & 0.9538 & 0.9503 \\
\midrule
\multirow{25}{*}{Criminal} & \multirow{5}{*}{Bodily injury} 
  & Polyglot-ko (1.3B) & 0.8792 & 0.8765 & 0.8778 \\
& & Exaone (2.4B)      & 0.8857 & 0.8953 & 0.8904 \\
& & \sys-360M    & 0.9306 & 0.9314 & 0.9310  \\
& & \sys-800M    & 0.9519 & 0.9515 & 0.9517  \\
& & \sys-1.5B    & 0.9518 & 0.9546 & 0.9532 \\
\cmidrule(lr){2-6}
& \multirow{5}{*}{Drunk driving} 
  & Polyglot-ko (1.3B) & 0.8697 & 0.8678 & 0.8688 \\
& & Exaone (2.4B)       & 0.8932 & 0.8894 & 0.8913 \\
& & \sys-360M    &0.8869 & 0.8832 & 0.8851 \\
& & \sys-800M    & 0.9045 & 0.9238 & 0.9140  \\
& & \sys-1.5B    & 0.9158 & 0.9231 & 0.9194 \\
\cmidrule(lr){2-6}
& \multirow{5}{*}{\makecell{Property theft\\and deception}} 
  & Polyglot-ko (1.3B) & 0.9097 & 0.9117 & 0.9107 \\
& & Exaone (2.4B)     & 0.8964 & 0.9010 & 0.8987 \\
& & \sys-360M   & 0.9212 & 0.9238 & 0.9225  \\
& & \sys-800M     & 0.9399 & 0.9396 & 0.9397\\
& & \sys-1.5B    & 0.9204 & 0.9246 & 0.9225 \\
\cmidrule(lr){2-6}
& \multirow{5}{*}{Sexual misconduct} 
  & Polyglot-ko (1.3B) & 0.8799 & 0.8759 & 0.877 \\
& & Exaone (2.4B)     & 0.8556 & 0.8441 & 0.8498 \\
& & \sys-360M   & 0.8888 & 0.8916 & 0.8902 \\
& & \sys-800M    & 0.9021 & 0.9045 & 0.9033 \\
& & \sys-1.5B     & 0.8768 & 0.8827 & 0.8797 \\
\cmidrule(lr){2-6}
& \multirow{5}{*}{Violence} 
  & Polyglot-ko (1.3B) & 0.8617 & 0.8669 & 0.8643  \\
& & Exaone (2.4B)      & 0.8679 & 0.8810 & 0.8744 \\
& & \sys-360M    & 0.8981 & 0.9118 & 0.9049 \\
& & \sys-800M    & 0.9043 & 0.9200 & 0.9120 \\
& & \sys-1.5B    & 0.9209 & 0.9416 & 0.9311 \\
\midrule
\multirow{5}{*}{Administrative} & \multirow{5}{*}{\makecell[l]{Administrative\\litigation}} 
  & Polyglot-ko (1.3B) &  0.8691 & 0.8654 & 0.8672 \\
& & Exaone (2.4B)      & 0.8913 & 0.8928 & 0.8920 \\
& & \sys-360M   & 0.9138 & 0.9097 & 0.9118 \\
& & \sys-800M     & 0.9372 & 0.9384 & 0.9377\\
& & \sys-1.5B    & 0.9297 & 0.9448 & 0.9372 \\
\bottomrule
\end{tabular}}

\caption{Token-level metrics (Precision, Recall, and Micro F1) for the \textbf{Per-Epoch Replacement} setting, reported by case type and model (parameters shown in parentheses).}

\label{tab:case-token-per-epoch}
\end{table*}

\section{License}

\end{document}